%% file: main.tex
\documentclass[10pt,twocolumn,letterpaper]{article}

\usepackage{wacv}
\usepackage{times}
\usepackage{epsfig}
\usepackage{graphicx}
\usepackage{amsmath}
\usepackage{amssymb}
\usepackage{multirow}
\usepackage{subcaption}
\usepackage{booktabs}
\usepackage[table]{xcolor}
\usepackage[breaklinks=true,bookmarks=false]{hyperref}

\graphicspath{ {./images_new/} }
\DeclareMathOperator*{\argmin}{argmin}
\newcommand{\norm}[1]{\left\lVert#1\right\rVert}

\wacvfinalcopy

\def \iccvsubmission {} 
\def \wacvsubmission {} 
\ifx \iccvsubmission \undefined
	\newcommand{\siva}[1]{\textcolor{brown}{Siva: #1}}
	\newcommand{\sm}[1]{\textcolor{brown}{#1}}
    \newcommand{\shivam}[1]{\textcolor{blue}{#1}}
    \newcommand{\sduggal}[1]{\textcolor{blue}{Shivam: #1}}
	
	\newcommand{\shenlong}[1]{\textcolor{magenta}{Shenlong: #1}}
	
	\newcommand{\raquel}[1]{\textcolor{red}{Raquel: #1}}
\else
	\newcommand{\siva}[1]{}
	\newcommand{\sm}[1]{{#1}}
    \newcommand{\shivam}[1]{{#1}}
    \newcommand{\sduggal}[1]{}
	
	\newcommand{\shenlong}[1]{}
	
	\newcommand{\raquel}[1]{}
\fi

\ifx \wacvsubmission \undefined
    \newcommand{\wacvshivam}[1]{\textcolor{blue}{#1}}
    \newcommand{\wacvsduggal}[1]{\textcolor{blue}{Shivam: #1}}
\else
    \newcommand{\wacvshivam}[1]{{#1}}
    \newcommand{\wacvsduggal}[1]{}
\fi

\input{tex_files/commands}

\begin{document}

\title{\vspace{-10mm} Mending Neural Implicit Modeling for 3D Vehicle Reconstruction in the Wild}
\author{Shivam Duggal$\thanks{Equal Contribution.}^{*1}$  \stepcounter{footnote}Zihao Wang\thanks{Work done as part of internship.} $^{*1}$ Wei-Chiu Ma$^{1,3}$ Sivabalan Manivasagam$^{1,2}$ \\ Justin Liang$^{1}$ Shenlong Wang$^{1,2}$ Raquel Urtasun$^{1,2}$\\
{$^1$Uber ATG, $^2$University of Toronto, $^3$Massachusetts Institute of Technology}
}

\maketitle
\thispagestyle{empty}

\input{tex_files/abstract}

\input{tex_files/introduction}

\input{tex_files/related}
\input{tex_files/background}

\input{tex_files/method}

\input{tex_files/experiments}

\input{tex_files/conclusion}


\newpage
\appendix
\onecolumn

\begin{center}
\Large \bf \par Supplementary Material \\ Mending Neural Implicit Modeling for 3D Vehicle Reconstruction in the Wild
\vskip .5em
\vspace*{12pt}
\end{center}

\maketitle
\thispagestyle{empty}

\input{supp/abstract}
\input{supp/experimental_details}
\input{supp/experimental_results}

\newpage
\clearpage
{\small
\bibliographystyle{ieee_fullname}
\bibliography{egbib}
}

\end{document}

%% file: tex_files/commands.tex
\newcommand{\arxiv}[1]{\vspace{0mm}}

\newcommand{\bX}{\mathbf{X}}

\newcommand{\bo}{\mathbf{o}}

\newcommand{\cL}{\mathcal{L}}

%% file: tex_files/abstract.tex
\begin{abstract}
Reconstructing \sm{high-quality}
3D objects from sparse, partial observations \sm{from a single view} is of crucial importance for various applications in computer vision, robotics, and graphics. 
While recent neural implicit modeling methods show promising results on synthetic or dense data, they perform poorly on \sm{sparse and noisy} real-world data.
We discover that the limitations \sm{of a popular neural implicit model} are due to
\wacvshivam{lack of robust shape priors and} lack of \shivam{proper} regularization.
In this work, \sm{we demonstrate high-quality in-the-wild shape reconstruction using:}
(i) a deep encoder as a robust-initializer of the shape latent-code; (ii) \wacvshivam{regularized} test-time optimization of the latent-code; (iii) a deep discriminator as a learned 
\wacvshivam{high-dimensional} 
shape prior;
(iv) \sm{a novel curriculum learning strategy that allows the model to \shivam{learn shape priors} on 
synthetic data and smoothly transfer them to sparse real world data.}
\shenlong{i.v are not informative. talk about "a novel curriculum strategy that allows to gradually training model from simulated, clean and dense data supervisely to real-world, noisy and partial observervations in unsupervised fashion...."}
Our approach 
better captures the global structure, 
performs well on occluded and sparse observations, and 
registers well with the ground-truth shape.
We demonstrate superior performance over state-of-the-art 3D object reconstruction methods on two real-world 
datasets.
\end{abstract}

%% file: tex_files/introduction.tex
\section{Introduction}
\label{sec:introduction}
\begin{figure}[t]
\scalebox{1}{
\begin{centering}
\includegraphics[width=0.98\linewidth]{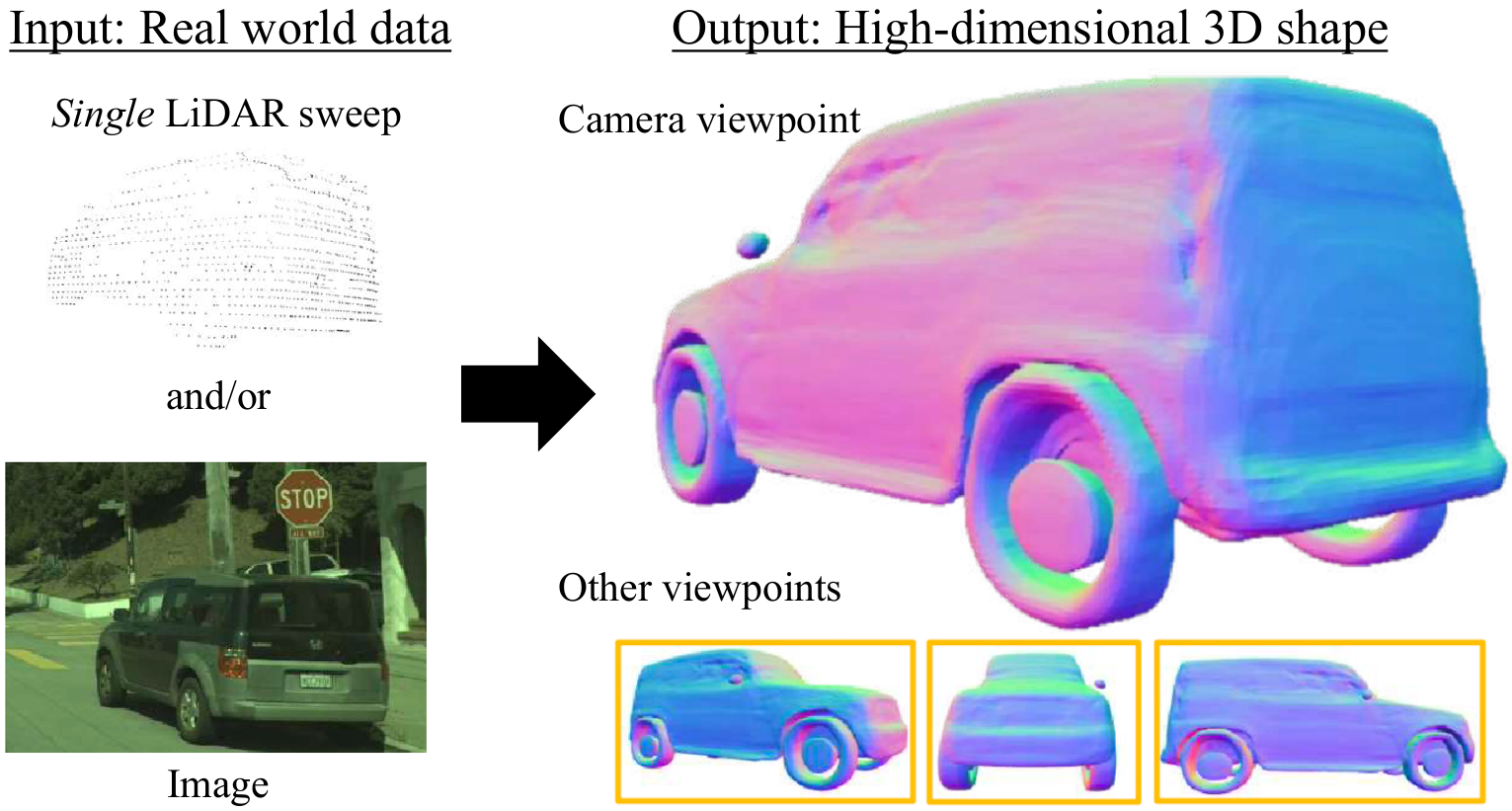}
\end{centering}}
\vspace{-4mm}
\caption{\textbf{3D Object Reconstruction in the Wild.} Our model takes as input a single LiDAR sweep \sm{and/or}
a RGB image, and outputs a high-quality 3D shape.}
\label{fig:teaser}
\vspace{-6mm}
\end{figure}

Consider the street view image and the partial LiDAR scan in Fig. \ref{fig:teaser}. As humans, we can effortlessly identify the complete vehicle in the scene and have a rough grasp of its 3D geometry. 
This is because human visual systems have accumulated hours of observations that help us develop mental models for these objects \cite{kar2015category,kanazawa2018learning}. While we may have never seen this particular car before, we know that cars shall be symmetric, they shall lie on the ground, and sedans shall have similar shapes and sizes. We can thus exploit these knowledge to infer the 3D structure of any object, given its sparse observations. The goal of this work is to equip computational visual machines with similar capabilities. One recently emerging class of such works is neural implicit shape modeling \cite{park2019deepsdf,ONet,mildenhall2020nerf,MetaSDF,sitzmann2019scene}. Neural implicit shape modeling involves reasoning about the surface of a 3D shape as the level-set of a function (represented using a neural network) \cite{Stan1991}, for example: an inside-outside function or a signed-distance function. 
In this work, we exploit the neural implicit shape representation for reconstructing high-\sm{quality} real-world objects. 

\begin{figure*}[h]
\begin{centering}
\includegraphics[width=0.98\linewidth]{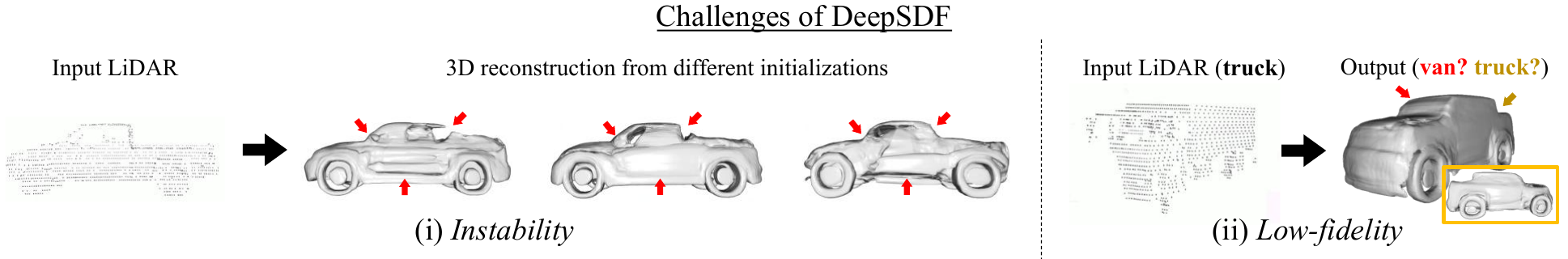}
\end{centering}
\caption{\textbf{Challenges of DeepSDF.} (i) \emph{Instability}: while the input observation remains intact, different initializations may lead to distinct 3D reconstructions. See the \textcolor{red}{red arrows}. (ii) \emph{Low-fidelity}: the model may easily overfit to the noise in the data, and fail to capture the global structure of the input observation. Here, the output shape is a hybrid of a van and a truck.}
\label{fig:challenges-deep-sdf}
\vspace{-5mm}
\end{figure*}

Existing neural implicit reconstruction methods operate by learning a low-dimensional mapping for the corresponding high-dimensional 3D shape. Such a low-dimensional mapping is usually represented as a 1D (or multi-dimensional) latent-code, which is implicitly decoded back to the 3D shape. Previous approaches usually take one of 
two routes to generate the low-dimensional latent-code: (i) learn a direct mapping from observations to the latent-code, or (ii) model this problem as a structured optimization task and incorporate human knowledge into the model. Specifically, former methods capitalize on machine learning algorithms to directly learn the statistical priors from data. While they are more robust to \sm{noise}
in the observation, the decoded high-dimensional 3D shapes are not guaranteed to match with the input observations. Additionally, many feed-forward methods \cite{ONet,chen2018implicit_decoder} are learned with ground-truth (GT) supervision from synthetic data, which is often unavailable for real-world data. This leads to poor generalization performance on unseen \sm{real-world} partial inputs. In contrast, optimization based approaches \cite{park2019deepsdf,dist} can produce coherent shape representations by incorporating the observations into carefully designed objective functions. In practice, however, designing the optimization objective function is cumbersome and 
hand-crafted priors \sm{alone cannot } 
include all possible phenomena. Despite hand-crafted regularization, optimization-based methods struggle with noise and \sm{can generate} 
unusual artifacts. These limitations 
\sm{are more pronounced } 
when reconstructing accurate and high-quality 3D shapes from real-world observations, because of 
potential noises, occlusions, and environment variability present in the real-world data. 

In this work, we address the shortcomings of the prior works by highlighting the key components necessary for reconstructing high-quality 3D shapes in the wild. 
\wacvshivam{In order to reconstruct high-fidelity shapes from sparse and noisy real-world observations, it is necessary for the proposed approach to leverage (1) \emph{strong shape priors} to extract the true 3D shape from input observations and be (2) \emph{robust to the variable noise} in the input. To satisfy these properties,}
we begin with using the deep encoder network (adopted from feed-forward approaches\cite{ONet,chen2018implicit_decoder}) as a \emph{robust initializer of the shape latent-code}. To guarantee that the predicted latent-code 
\sm{is faithful to}
the input observations, we further incorporate the \emph{test-time optimization of the latent-code}\cite{park2019deepsdf} into our shape reconstruction method. 
However, we still have not addressed the hand-crafted objective function limitation. While it is necessary to carefully design an objective function for effective training and inference of the shape reconstruction \sm{system,}
 the overall process is cumbersome and hand-designed. \wacvshivam{Moreover, test-time optimization is generally susceptible to noise in the observations.} To overcome these limitations, we introduce a learned shape prior into our objective function, which serves as a \emph{high-dimensional structural shape regularizer}. We do so by taking inspiration from the recent GAN inversion theory for 2D image generation \cite{zhu2020indomain,gu2020image} 
  and incorporate a 3D shape discriminator network into our \sm{approach.}
The discriminator judges the ``naturalness'' of the reconstructed shape both during training and test time. Finally, to effectively train our \sm{full model}
we introduce an \emph{adversarial curriculum learning strategy}. \sm{This allows us}
to effectively learn several shape priors from synthetic data, and generalize them to real-world data.

We evaluated our approach for 3D vehicle reconstruction in the wild on two challenging self-driving datasets. We compare against state-of-the-art models on three tasks: LiDAR-based shape completion, image-based shape reconstruction and image + LiDAR shape completion. Our results show superior performance in terms of both reconstruction accuracy and visual quality.
\wacvshivam{Overall, we tackled the less-explored task of high-fidelity reconstruction of diverse real-world vehicles. We showcased significant improvement compared to prior works, thanks to the learned robust shape priors, the curriculum learning strategy used to extract them and the regularized optimization strategy.}

\shenlong{one sentence talk about sim2real without real data performance: "importantly, with the help of xx, our method produce... without any form supervision from real-world"}
\siva{"without any form of supervision" sounded too strong for me given experiments.}

%% file: tex_files/related.tex

\begin{figure*}
\begin{center}
\includegraphics[width=0.9\linewidth]{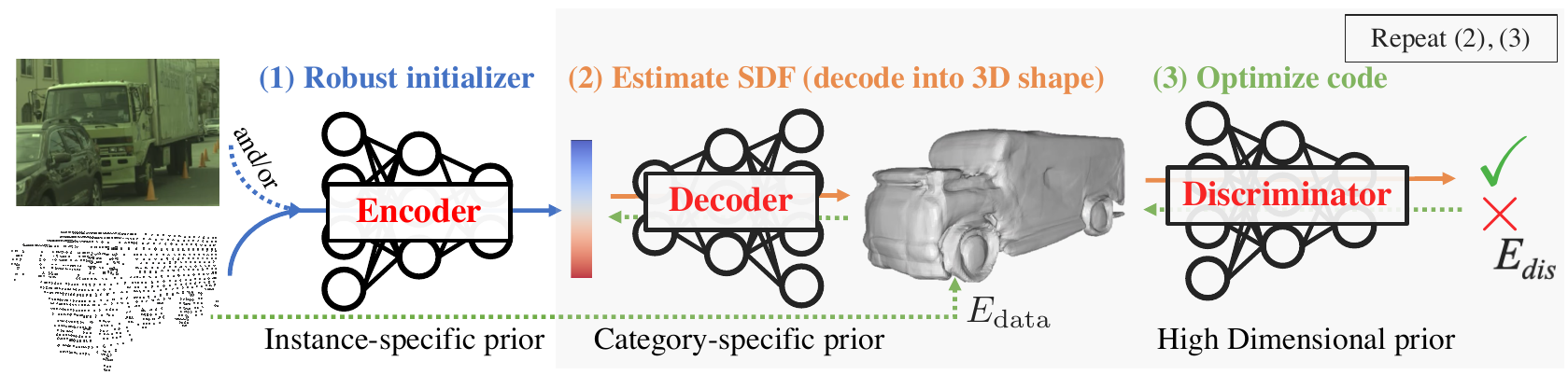}
\vspace{-5mm}
\end{center}
\caption{\textbf{Method Overview:} 
Given raw sensory data, our method first exploits a deep encoder to predict a robust initialization for the shape code. The latent shape-code is then optimized through the auto-decoder framework using a data-consistency energy term $(E_\mathrm{data})$ and a discriminator-induced high-dimensional shape regularization term $(E_\mathrm{dis})$.
}
\label{fig:architecture}
\vspace{-4mm}
\end{figure*}

\section{Related Work}

\paragraph{Feed Forward Shape Completion:}
Deep feed-forward
approaches encode images or depth \sm{sensor data}
into latent representations, which are subsequently decoded into complete meshes or point clouds. 
Early works used voxel-grid based representations \cite{dai2017shape, han2017high} which had lossy resolution and memory constraints. 
Other works instead leveraged point clouds directly to create latent codes \cite{qi2017pointnet, qi2017pointnet++,gu2020weaklysupervised} for point cloud completion or mesh construction, using a variety of decoder architectures \cite{yang2018foldingnet, yuan2018pcn, liu2020morphing, tchapmi2019topnet, wen2020point}, enhanced intermediate representations \cite{xie2020grnet, groueix2018atlasnet}, and GAN-based loss functions \cite{sarmad2019rl}. 
While 
successful on synthetic datasets \cite{chang2015shapenet} or dense 3D scans \cite{dai2017scannet}, 
\wacvshivam{prior works \cite{yuan2018pcn, xie2020grnet,gu2020weaklysupervised} have had limited success on real world noisy datasets such as KITTI \cite{geiger2012we}}.
Other works \cite{gkioxari2020mesh, dai2019scan2mesh} directly perform mesh prediction from images and 3D data. 
Recent works \cite{chen2018implicit_decoder,ConvolutionOccNw, ONet, PiFU, DISN,saito2020pifuhd,trevithick2020grf,niemeyer2020differentiable} represent \shivam{shapes using neural implicit functions}
and have shown promising results on synthetic datasets \cite{chang2015shapenet, RenderPeople} or dense scans \cite{chang2017matterport3d}, but few have shown large-scale results on real data \cite{gu2020weaklysupervised}. \cite{Stutz_2018, najibi2020dops} 
\shivam{adapted feed-forward networks} trained on synthetic data to real data by training an encoder that 
\wacvshivam{transforms}
partial point clouds 
\wacvshivam{to}
latent codes that match 
the 
observations. \wacvshivam{Concurrently, \cite{Bechtold2021HPN} showcased the importance of hierarchical priors (local and global shape priors) in generalizing to (unseen) real-world shapes.} However, to our knowledge, most previous 
works applied to real sparse data generate shapes that are often amorphous, overly smooth, or restricted to \sm{small vehicles.}

\vspace{-4mm}
\paragraph{Optimization-based Shape Completion:}
Another line of work frames shape completion as an optimization problem, where the objective is to ensure the predicted shape is consistent with real sensor observations. Such works require strong shape priors, as the \shivam{real-world observations} are quite sparse, noisy, and have large holes.
\cite{Engelmann16GCPR, Engelmann_2017, wang2020directshape, 3DRCNN_CVPR18} represent the shape prior as a PCA embedding of the volumetric signed distance fields, and optimize the shape and pose given sensor
data. While showing promising performance on pose estimation and object tracking, the recovered shapes are coarse due to low dimensional linear embeddings, and \shivam{generally represent sedans or smaller vehicles only.}
\shivam{Recent works encode shape priors via neural implicit representations, and perform 
optimization supervised by input 3D observations \cite{park2019deepsdf,davies2021effectiveness}, with some using differentiable rendering
~\cite{dist,  jiang2020sdfdiff, lin2020sdfsrn}.} While \cite{park2019deepsdf,dist,jiang2020sdfdiff} optimized the low-dimensional latent-code, \cite{lin2020sdfsrn,davies2021effectiveness,YangWenCVPR21} directly optimized the neural network weights per object. Most of these works require dense supervision from the whole 3D space
(GT occupancy or signed distance values) during training, preventing them \sm{from training} 
on sparse real-world data. 
\shivam{Thus, these works have focused only on synthetic shape completion.}
\cite{zakharov2020autolabeling} have explored this approach for auto-labeling, but do not focus on \sm{shape quality}.

\vspace{-6mm}
\paragraph{Latent Space Optimization:} Optimizing network weights or latent codes at test time has been quite popular for several tasks \cite{bloesch2019codeslam, chen2020category, zhu2020indomain, park2019deepsdf}. 
CodeSLAM \cite{bloesch2019codeslam} optimizes geometric and motion codes during inference time for VisualSLAM, Chen~\etal\cite{chen2020category} optimizes geometric and appearance codes at inference time for pose estimation.
Recently, latent code optimization has been explored in the field of GAN inversion \cite{zhu2020indomain, shen2020interfacegan, gu2020image}  \wacvshivam{for 2D image generation.}
In this work, we take inspiration from the theory of GAN inversion and apply it to the task of high-fidelity 3D reconstruction.
Another work related to ours is \cite{gurumurthy2019high},  which performs latent code optimization for point cloud completion and uses a GAN to ensure the optimized latent code is in the generator's manifold. \wacvshivam{However, they demonstrated their approach on a controlled synthetic data environment and didn't showcase any generalization to unseen data. Moreover, unlike our discriminator, their discriminator acts on a low-dimensional space. By directly passing the high-dimensional 3D shape to the discriminator, we allow the gradients to directly backpropagate from the reconstructed shape to the latent-code and are thus able to guide the optimization to generate realistic shapes.}

%% file: tex_files/background.tex
\section{Background}
\label{sec:background}
The goal of 3D object shape reconstruction is to recover 3D geometry of an object given sensor observations (e.g: point cloud, images, etc). 
In this work, we parameterize a 3D shape implicitly using a signed-distance function, i.e. given a 3D point $\mathbf{x}$, we model the function $f(\mathbf{x}) = \mathbf{s}$, where $\lvert \mathbf{s} \rvert$ equals the distance of point $\mathbf{x}$ from the surface, and $\mathit{sign}(\mathbf{s})$ represents inside / outside status of point $\mathbf{x}$ w.r.t. the 3D object.
\sm{We then represent shape reconstruction as:}
\[f_{\theta}(\mathbf{x}, g_{\phi}(\mathbf{o}))  = \mathbf{s},  \vspace{-2mm}\] where function $g$ encodes the input observations $\mathbf{o}$ into some latent representation. $\theta$ and $\phi$ are the parameters of the functions $f$ and $g$ respectively. 
\sm{To extract the final mesh, one can query} the SDF values for a \sm{grid of points in the volume and then perform }
marching cubes \cite{MarchingCubes}.

Recent neural implicit modeling methods \cite{park2019deepsdf,ONet,chen2018implicit_decoder} generate high-fidelity shapes on synthetic or dense datasets. However, their shape \sm{quality}
degrades significantly on sparse and noisy real-world datasets. 
Towards our goal of reconstructing \sm{high quality}
shapes on real-world \sm{data,}
we first carefully analyze 
\sm{two standard approaches \shivam{which} prior works follow}
and highlight their limitations. 

\begin{figure*}
\centering
\includegraphics[width=0.90\linewidth]{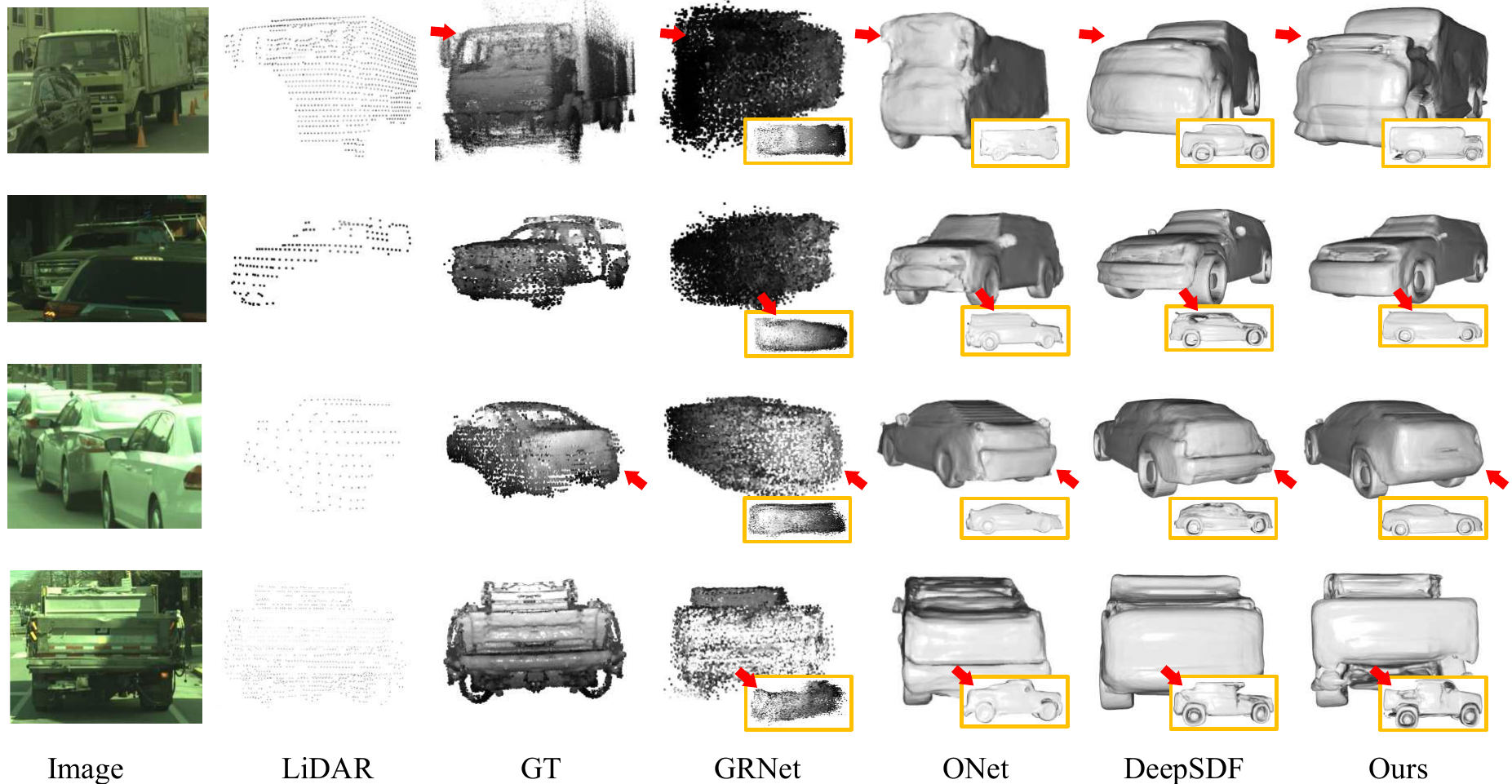}
\caption{\textbf{LiDAR Completion on NorthAmerica: }Compared to baselines, our approach: (Row 1) captures the global structure, (Row 2) performs well in occluded \shivam{scenarios}, (Row 3) registers well with \sm{both input and GT points,}
(Row 4) maintains finer details (see cavity at the back of the car). Check supp. for additional comparisons 
\sm{(such as with Ours$_\text{no-finetune}$).}}
\label{fig:qual-NorthAmerica}
\vspace{-4mm}
\end{figure*}

\subsection{Feed-forward Approaches} \label{sec:feed-forward}Feed-forward networks \cite{ONet,chen2018implicit_decoder} formulate the above shape reconstruction task as an encoder-decoder task. The goal of the encoder ($g_{\phi}$) is to extract discriminative latent cues, $\mathbf{z}$, from input observations (\shivam{$\mathbf{o}$}), and the role of the decoder ($f_{\theta}$) is to map a 3D point (conditioned on the encoded latent features $\mathbf{z}$) to its corresponding signed-distance value.

During training, the encoder and decoder are jointly optimized \sm{with the}
GT signed distance field as supervision. 
The GT is basically a set of $(\mathbf{x}, \mathbf{s})$ pairs, where the 3D points span over the whole volume (generally a normalized cube, enclosing the GT object). 
By generalizing the encoder over the training \shivam{instances,
feed-forward approaches make the encoder robust to \sm{input}
 noise.}
However, this \sm{can cause }
over-smooth shapes at test-time, which \sm{are not realistic and} do not register well 
\sm{with the observations} \shivam{(See ONet\cite{ONet} results in Fig.~\ref{fig:qual-NorthAmerica})}. 
\sm{Low-quality shape predictions and artifacts are particularly pronounced for out-of-distribution inputs.}

\subsection{Neural Optimization Approaches} This line of neural implicit 
reconstruction works \cite{park2019deepsdf,dist,davies2021effectiveness,MetaSDF,Stutz_2018} focuses on combining the representation power of neural networks with structured optimization. 
One \sm{example}
is DeepSDF \cite{park2019deepsdf}, which proposed an ``auto-decoder'' based optimization approach. Instead of inferring the latent code from observations $(\mathbf{z} = g_{\phi}(\mathbf{o}))$, the auto-decoder framework directly optimizes the latent code, such that the predicted shape registers with the input observations.

During training, the DeepSDF model learns to both decode the spatial point $\mathbf{x}$ into SDF value $\mathbf{s}$ (via the decoder $f_{\theta}$) and assign a latent code $\mathbf{z}$ to each training shape. 
The latent code learns to be a compact yet expressive representation of the 3D shape. Such a training procedure leads to the decoder emphasizing a \emph{category-specific shape prior} (by generalizing over all training instances of a category) and the latent code encoding an \emph{instance-specific shape prior}. 
\sm{During inference,} the decoder is kept fixed, and the latent code \shivam{(randomly initialized from a normal distribution)} is optimized to minimize the following energy function:
\begin{align}
\mathbf{z}^\ast = \argmin_\mathbf{z} E_\mathrm{data}(\mathbf{o}, f_\theta(\mathbf{X}, \mathbf{z})) + \lambda E_\mathrm{reg}(\mathbf{z}).
\label{eq:sdf-optim}
\end{align}
where $\mathbf{X}$ \sm{typically} represents a sparse set of input 3D points. 
The data term $E_\mathrm{data}$ ensures consistency between the estimated 3D shape and the observation, 
while the regularization term $E_\mathrm{reg}$ constrains the latent code. The final shape can be obtained by querying the SDF value (via $f_\theta(\mathbf{x}, \mathbf{z}^*)$).

In practice, 
 for shape completion, DeepSDF requires 
GT $({\mathbf{x}, \mathbf{s}})$ pairs\footnote{\wacvshivam{During inference, DeepSDF not only uses on-surface partial point cloud (SDF=0), but also non-surface points with known GT SDF values.}}
The data term is implemented as clamped $\mathcal L_{1}$ distance between estimated SDF value $f_{\theta}(\mathbf{x}, \mathbf{z})$ and the corresponding GT SDF value, $\mathbf{s}$. The regularization term equals $\ell_2$-norm of the latent code. 

\begin{figure*}
\centering
\includegraphics[width=0.85\linewidth]{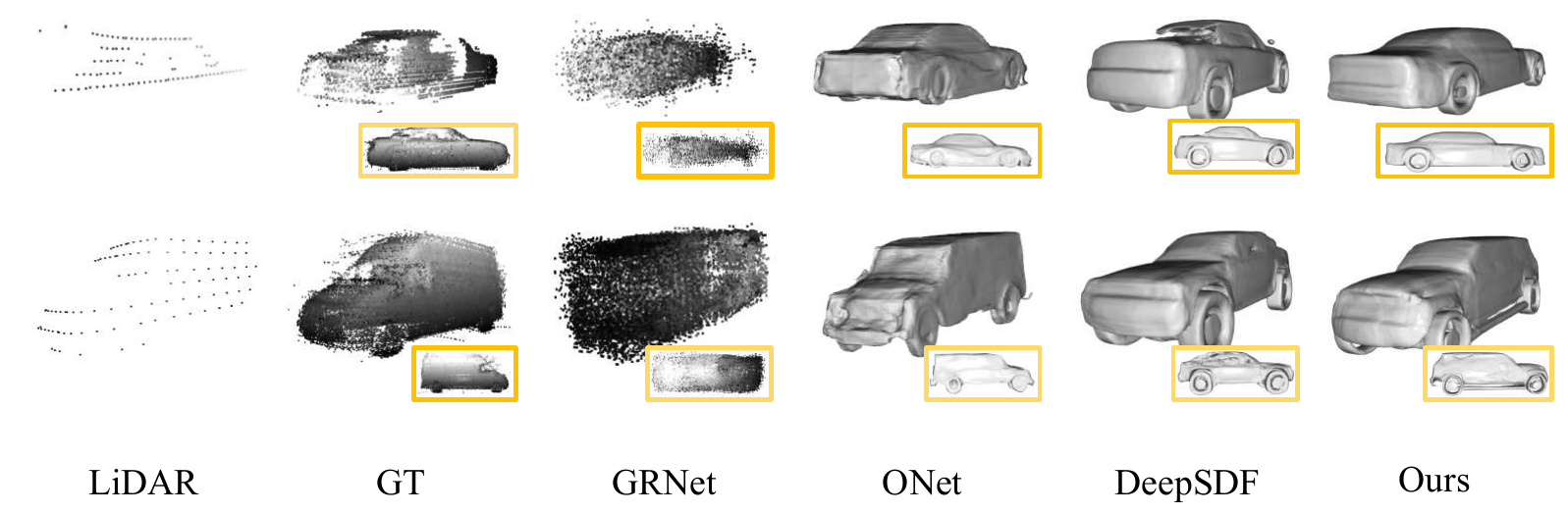}
\vspace{-2mm}
\caption{\textbf{LiDAR Completion on KITTI:} Compared to others, our approach maintains high fidelity with input points and generates finer shape details even with sparse input data. However, when input is significantly sparse, fidelity with GT drops.}
\label{fig:qual-KITTI}
\vspace{-5mm}
\end{figure*}

\vspace{-5mm}
\paragraph{Limitations:} \textbf{(1) Instability}: 
The 
 auto-decoder framework can suffer from \sm{shape instability w.r.t.}
different latent code initializations.
\sm{This is because the optimization landscape for Eq. \ref{eq:sdf-optim} is highly non-convex.}
 As shown in Fig.~\ref{fig:challenges-deep-sdf} (i), a minor perturbation in initialization may lead to completely different local \shivam{minima} and hence the final output. 
\sm{\textbf{(2) Low-quality}:}
The optimal code is not guaranteed to generate a \sm{high-quality}
and ``natural'' shape.
This limitation (Fig.~\ref{fig:challenges-deep-sdf} (ii)) is due to lack of structural regularization in the objective. 
Specifically, the data term $E_\mathrm{data}(\bo, f_{\theta}(\mathbf{}{X}, \mathbf{z}))$ in Eq. \ref{eq:sdf-optim} is typically decomposed into sum of independent terms per each individual point $\sum_{\mathbf{x} \in \mathbf{X}} E_\mathrm{data}(\mathbf{o}, f_\theta(\mathbf{x}, \mathbf{z}))$.  There is little constraint on the global SDF field --- each data point operates individually. The model can thus easily overfit to noise it has never seen during training and the latent code may drop out of manifold during optimization, leading to artifacts. 
\shivam{This issue is particularly severe for real-world scenarios, where observations are sparse and noisy.}

%% file: tex_files/method.tex

\section{Method}
\label{sec:method}

\sm{We now present our method for reconstructing accurate shapes from real-world observations.
 To address previous works' limitations \wacvshivam{on real-world datasets}, we combine components from both feed-forward and neural optimization approaches.
 From the feed-forward \shivam{approaches,}
 we leverage an encoder \shivam{to robustly initialize a latent-code, given partial real world observations.}
 Following neural-optimization, \shivam{we utilize the auto-decoder framework to further optimize the latent-code, ensuring higher shape fidelity.}
 However, the key challenge for shape reconstruction in the wild is that given \wacvshivam{(noisy, sparse)} out-of-distribution observations,} \wacvshivam{the reconstructed shapes no longer remain within the domain of realistic and naturally-looking shapes. To ensure that our reconstructed shapes always look realistic and maintain high-level properties of the shape category, we incorporate a high-dimensional shape regularizer in form of a discriminator, at both training and test-time optimization.}

\sm{Thus,} our
shape reconstruction pipeline consists of encoder, decoder and discriminator
as shown in Fig.~\ref{fig:architecture}. 
\sm{\wacvshivam{Finally,} to ensure proper learning of all components,} we propose a \wacvshivam{novel} adversarial curriculum learning strategy. 
\sm{We now describe in detail our discriminator (Sec. \ref{sec:discriminator}), modified 
\wacvshivam{test-time optimization}
procedure (Sec. \ref{sec:inference}), and curriculum learning strategy (Sec. \ref{sec:curriculum-learning}).}

\subsection{Discriminator as Shape Prior} 
\label{sec:discriminator}
Based on the observation that neural optimization based approaches lack a strong global shape prior, we introduce a discriminator to induce a learned 
prior over the \sm{predicted} SDF. 
 The discriminator function $D_\psi$ (parameterized by $\psi$) evaluates the likelihood \sm{that the reconstructed shape is ``natural'' relative to a set of ground-truth synthetic shapes (e.g., ShapeNet).}
 The benefits of a discriminator are two-fold: (1) During training, it serves as an objective function to improve the \sm{encoder} 
 through adversarial loss. (2) During inference, \sm{it regularizes the optimization process to ensure the latent code remains in the decoder's domain.}
 To capture the overall geometry of the SDF,
  we randomly sample $N$ points from the whole 3D space, query their SDF value, and pass all of them into the discriminator, which outputs a single scalar representing the ``naturalness'' of the shape. 

\subsection{Regularized Optimization for Real-World \\Inference} 
\label{sec:inference}
At inference time, given sparse and noisy real-world observations $(\mathbf{o})$, \sm{an encoder}
predicts a robust initialization for the shape latent code ($\mathbf{z} = g_{\phi}(\mathbf{o})$). Then, to 
ensure that the predicted shape aligns well with the observations, we optimize the latent code using the following objective function:
\begin{align}
\begin{split}
\mathbf{z}^\ast = \argmin_\mathbf{z} E_\mathrm{data}(\mathbf{o}, f_{\theta}(\mathbf{X}, \mathbf{z})) + \lambda_\text{reg} E_\mathrm{reg}(\mathbf{z}) \\ + \lambda_\text{dis}E_\mathrm{dis}(f_{\theta}(\mathbf{X}, \mathbf{z})).
\end{split}
\label{eq:domain-optim}
\end{align}
Our data term, $E_\mathrm{data}$, and latent-code regularization term, $E_\mathrm{reg}(\mathbf{z})$, have the same format as DeepSDF (Eq.~\ref{eq:sdf-optim}):
\[E_\mathrm{data}(\mathbf{o}, f_{\theta}(\mathbf{X}, \mathbf{z}))  = \sum_{x \in X} \rho( \mathbf{s}, f_\theta(\mathbf{x}, \mathbf{z}) ) \vspace{-3mm}\]
where $\mathbf{s}$ is the actual signed distance value at point $\mathbf{x}$, computed from the observation $\mathbf{o}$. $\rho$ is a robust clamped $\mathcal L_{1}$ distance. The regularization term is the $\ell_2$-norm of the latent code (\ie, $E_\mathrm{reg}(\mathbf{z})  = \| \mathbf{z} \|_2^2$). \shivam{The additional discriminator prior term, $E_\mathrm{dis}(f_{\theta}(\mathbf{X}, \mathbf{z})) = -\log D_{\psi}(\mathbf{X}, f_{\theta}(\mathbf{X}, \mathbf{z}))$, encodes the belief of the discriminator that the generated shape is \sm{natural}.}
The 3D points ($\mathbf{X}$) used for $E_\mathrm{dis}$ are sampled from the whole 3D space, while the 3D points for $E_\mathrm{data}$ come directly from the observations. 

\begin{table}[t]
\centering
\scalebox{0.95}{
\begin{tabular}{lcccc}
\specialrule{.2em}{.1em}{.1em}
Method & ACD (mm) $\downarrow$ & Recall (\%) $\uparrow$\\
\specialrule{.1em}{.05em}{.05em}
ONet \cite{ONet} & 22.76 & 49.56\\
GRNet \cite{xie2020grnet} & 12.70 & 77.59\\ \hline
SAMP \cite{Engelmann_2017} & 176.42 & 65.58\\
DIST \cite{dist} & 19.55 & 71.54\\
DIST++ \cite{dist} & 17.29 & 72.50\\
DeepSDF \cite{park2019deepsdf} & 8.34 & 84.71\\
Ours$_\text{no-finetune}$ & 7.02 & 86.48\\
Ours$_\text{finetune}$ & \textbf{5.93} & \textbf{88.18}\\
\specialrule{.1em}{.05em}{.05em}
\end{tabular}
}
\vspace{-2mm}
\caption{\textbf{LiDAR Completion Results on NorthAmerica.}}
\label{tab:quan-na}
\vspace{-4mm}
\end{table}

\subsection{Adversarial Curriculum Learning Strategy} 
\label{sec:curriculum-learning}
\shivam{Training neural implicit models require large amounts of dense 3D ground-truth data.} \sm{While synthetic data provides this dense supervision and helps the model gain a rich prior over shapes, real-world in-the-wild data lacks dense GT.} \shivam{To effectively transfer the model trained on synthetic data to real-world data, its important to ensure that each component encodes complementary rich shape priors.}
\sm{Simply jointly training the full model on synthetic data and directly applying to real data results in poor performance. We therefore introduce a curriculum learning strategy that allows our model to generalize to real-world data.}
\shenlong{2-3 sentence to motivate why you need this: 1) dense supervision is not available in real-world; 2) sim provides rich prior information about how output looks but not realistic 3) directly train on sim and apply on real is challenge, thus we need a curriculum to guide such transfer. Maybe Siva can help? }
We \sm{split}
training
into two stages: Stage 1 and Stage 2. In Stage 1, we train the DeepSDF-based auto-decoder framework on clean and dense synthetic data. Then, to ensure our approach can handle sparse inputs at test time, in Stage 2 we perform adversarial training of the encoder using sparse synthetic observations. This additional Stage 2 allows using (dense and clean) synthetic GT data as strong supervision, while training the encoder to encode sparse observations.

\vspace{-3mm}
\paragraph{Stage 1:} 
We first train the decoder over synthetic training data, in the same manner as DeepSDF. Specifically, we jointly optimize the decoder weights and latent code to reconstruct the ground truth (GT) signed distances:
\[\min_{\{\mathbf{z}_i\}, \theta}  \cL^\text{dec} \;+\; \cL^\text{reg} \vspace{-3mm}\]
\sm{where $\cL^\text{dec} = \min_{\{\mathbf{z}_i\}, \theta}  \sum_i^M \sum_j^N \rho(\mathbf{s}_{i,j}, f_{\theta}(\mathbf{x}_{i, j}, \mathbf{z}_i))$ and $\cL^\text{reg} = \lambda_{\mathrm{reg}}  \| \mathbf{z}_i \|_2^2$.}
$M$ is the total number of training shapes and $N$ is the number of training point samples per shape. $\mathbf{s}_{i, j}$ is GT signed distance value for a training point sample $\mathbf{x}_{i, j}$. 
$\rho$ refers to clamped $\mathcal L_{1}$ distance. 
Once the decoder ($f_{\theta}$) is trained, we keep it fixed, preserving its ability to generate ``natural" shapes. 
The optimized latent codes \sm{$(\mathbf{z}')$} and the predicted signed distance fields \shivam{$(\mathbf{s}' = f_{\theta}(\mathbf{x}, \mathbf{z}'))$} are used as pseudo-ground-truth (pseudo-GT) in the next stage.

\vspace{-3mm}
\paragraph{Stage 2:} \sm{Next, given a synthetic \textit{partial} point cloud or image observation, we adversarially train an encoder to predict a shape latent code $(\mathbf{z} = g_{\phi}(\mathbf{o}))$ in the pre-trained decoder's domain.}
\sm{For each synthetic training shape, we want to ensure 
that the predicted signed distance values match \shivam{well with GT values}
$(\cL^\text{dec})$, and that
the encoder-predicted latent code 
matches the pseudo-GT latent code 
from Stage 1 $(\cL^\text{z})$.
In addition, we simultaneously train a discriminator to ensure the predicted signed-distance field matches the ``naturalness'' of the pseudo-GT signed distance field from Stage 1 $(\cL^\text{gan})$\footnote{\shivam{We observed that using Stage 1 predicted SDF ($\mathbf{s}'$) instead of GT SDF as real samples for GAN loss results in better training and better results.}}.}
The overall loss function for training is:
\vspace{2mm}
\[\cL  = \cL^\text{dec} + \cL^\text{z} + \cL^\text{gan} \vspace{-2mm} \]
$\cL^\text{dec}$ measures the clamped  $\mathcal L_1$ distance between the estimated \shivam{and GT SDF }
values for each training point sample (same as $\cL^\text{dec}$ loss in Stage 1). 
$\cL^\text{z}$ measures the $\mathcal L_2$ distance between the encoder-predicted code and the pseudo-GT latent code 
($\cL^\text{z} = \| \mathbf{z} - \mathbf{z'}\|$). \wacvshivam{Additionally, we also enforce the encoder to embed the input sparse data into a realistic shape's latent code (using $\mathbf{- \sum_{x \in X} \log(D_{\psi}(\mathbf{x}, f_{\theta}(\mathbf{x}, \mathbf{z})))}$ as loss)}.
$\cL^\text{gan}$ is the GAN loss between predicted SDF field and pseudo-GT SDF field \wacvshivam{for training the discriminator}:
 \[\cL^\text{gan}  = \sum_{\mathbf{x} \in \mathbf{X}} \log(D_{\psi}(\mathbf{x}, f_{\theta}(\mathbf{x}, \mathbf{z}))) \;+\; \log(1-D_{\psi}(\mathbf{x}, 
 \mathbf{s}')) \vspace{-2mm}
 \] 
Such a training procedure disentangles the learning of the encoder, decoder and discriminator module. Through stage 1, the decoder learns to induce \emph{category-specific shape prior}, by generalizing over all instances of the training set. By keeping the decoder fixed and training the encoder to generate an instance specific latent code, the encoder learns to induce \emph{instance-specific shape prior}, while the discriminator acts as a \emph{high-dimensional structural shape prior}. 

%% file: tex_files/experiments.tex
\section{Experimental Details}
\label{sec:experiments}

\begin{table}[t]
\centering
\scalebox{0.85}{
\begin{tabular}{lcccc}
\specialrule{.2em}{.1em}{.1em}
Method & ACD (mm) $\downarrow$ & Recall (\%) $\uparrow$\\
\specialrule{.1em}{.05em}{.05em}
ONet \cite{ONet} & 18.10 & 59.10\\
GRNet \cite{xie2020grnet} & 13.66 & 78.21\\ \hline
DeepSDF \cite{park2019deepsdf} & 14.81 & 79.76 \\ 
Ours$_\text{no-finetune}$ & \textbf{8.60} & \textbf{82.97}\\
\specialrule{.1em}{.05em}{.05em}
\end{tabular}
}
\vspace{-2mm}
\caption{\textbf{LiDAR Completion Results on KITTI.}}
\label{tab:quan-kitti}
\vspace{-4mm}
\end{table}

\subsection{Datasets}

\paragraph{ShapeNet:} We exploit 2364 watertight cars from ShapeNet~\cite{chang2015shapenet} as our synthetic \sm{training} dataset. We follow DeepSDF \cite{park2019deepsdf} to generate the GT SDF samples for the first stage of training. For stage 2, we simulate image and sparse point clouds (as input observations) for each object from five different viewpoints, resulting in a total dataset of 11820 $(2364 \times 5)$ image-point cloud pairs. 

\vspace{-4mm}
\paragraph{NorthAmerica:} We {further} build a novel, large-scale 3D vehicle reconstruction dataset using a self-driving platform that collects data over multiple metropolitan cities in North America {and} under various weather conditions and time of the day. 
Using the acquired data, we generate a test-set of {935} high-quality instances. \shivam{Each consists of aggregated LiDAR sweeps used as GT shape, partial single LiDAR sweep and cropped camera image used as input observations.}

\paragraph{KITTI:} We also generate 209 diverse \shivam{test-set} objects from 21 sequences of the KITTI tracking dataset~\cite{geiger2012we}. For each object, we construct a dense ground truth 3D shape by aggregating multiple \shivam{LiDAR} sweeps using the GT bounding boxes. 
\shivam{KITTI dataset is used only during test time for evaluation. Please refer to supp. for more dataset details.}

\begin{table}[t]
\centering
\begin{tabular}{lcccc}
\specialrule{.2em}{.1em}{.1em}
Method & ACD (mm) $\downarrow$ & Recall (\%) $\uparrow$\\
\specialrule{.1em}{.05em}{.05em}
DIST~\cite{dist}  & 62.97 & 48.82\\
Ours & \textbf{8.89} & \textbf{84.32}\\
\specialrule{.1em}{.05em}{.05em}
\end{tabular}
\vspace{-2mm}
\caption{\textbf{Single Image Reconstruction on NorthAmerica}} 
\label{tab:im}
\vspace{-4mm}
\end{table}

\subsection{Metrics}Unlike their synthetic counterpart, real-world datasets do not possess complete watertight 3D shapes. We cannot evaluate 3D reconstruction metrics like Chamfer Distance, Volumetric IoU \cite{riegler2017octnet}, normal consistency \cite{riegler2017octnet}, and robust F-score \cite {Tatarchenko2019SingleImage} {on them}. We thus adopt asymmetric Chamfer Distance (ACD) between the GT point cloud and the reconstructed shape to measure the shape fidelity. ACD is defined as the sum of squared distance of each ground truth 3D point {to the closest surface point on} the reconstructed shape:

\[\mathrm{ACD}(\bX, \mathbf{Y}) = \frac{1}{|\bX|}\sum_{x \in \bX} \min_{y \in \mathbf{Y}} \norm{x-y}_{2}. \vspace{-2mm}\] 

We also compute the recall of the ground truth points from the reconstructed shape as a robust alternative:
\[\mathrm{Recall(\bX, \mathbf{Y})} = \frac{1}{|\bX|}\sum_{x \in \bX} \bigg[ \min_{y \in\mathbf{Y}} \norm{x-y}_{2} <= t \bigg]. \vspace{-2mm}\]

We set true-positive threshold  $t=0.1$ m in the paper. \shivam{All} the reported metrics 
are averaged over the test set objects.

\subsection{Baselines}\shivam{We compare against several state-of-the-art 3D object reconstruction algorithms:}
(i) feed-forward methods, such as Occupancy Networks (ONet) \cite{ONet} and GRNet \cite{xie2020grnet}; (ii) (deep) optimization based methods, such as {linear SAMP \cite{Engelmann_2017}}, DeepSDF \cite{park2019deepsdf}, and DIST \cite{dist}. {DeepSDF} samples off-surface points for SDF optimization to improve robustness. We also augment the original DIST approach with such sampling procedure, referred to as DIST++. 

\section{Experimental Results}

We now showcase our experimental results. We first compare our proposed approach against \shivam{the baselines}
on the LiDAR-based 3D shape completion task (Sec. \ref{sec:exp_lidar}). Next, we extend our approach to other sensor modalities (images \sm{(Sec. \ref{sec:exp_image})} and image + LiDAR combined \sm{(Sec. \ref{sec:exp_lidar_image})}). 
Finally, \sm{we perform an ablation of our approach (Sec. \ref{sec:exp_ablation}).}

\subsection{LiDAR-based Shape Completion} 
\label{sec:exp_lidar}
LiDAR-based shape completion refers to the task of reconstructing a 3D shape, given its partial point-cloud as the input observation ($\mathbf{o}$). 
We report LiDAR-based shape reconstruction results on NorthAmerica and KITTI datasets in Tab.~\ref{tab:quan-na} and Tab.~\ref{tab:quan-kitti} respectively. ``Ours$_{\text{no-finetune}}$'' result was generated by solely training the proposed approach on the 
ShapeNet dataset (as mentioned in Sec.~\ref{sec:curriculum-learning}), followed by inference-stage optimization. Ours$_{\text{no-finetune}}$ approach significantly outperforms all the baselines. In particular, we reduce ACD error by 45\% (37\%) compared to the best feed-forward network GRNet and 16\% (42\%) compared to the
best optimization method DeepSDF on NorthAmerica (KITTI) dataset. \shivam{In contrast to NorthAmerica,
GRNet performs better than optimization-based DeepSDF on KITTI.}
As KITTI point clouds are noisier, this demonstrates the value of feed-forward approaches being robust. We note that SAMP has particularly large error on NorthAmerica dataset due to its difficulty in handling larger objects of the dataset (such as trucks), as the embedding was trained mostly with smaller synthetic vehicles. Overall, the improvements suggest the effectiveness of our approach compared to feed-forward or optimization-only methods.

\begin{table}[t]
\centering
\begin{tabular}{lcccc}
\specialrule{.2em}{.1em}{.1em}
Method & ACD (mm) $\downarrow$ & Recall (\%) $\uparrow$\\
\specialrule{.1em}{.05em}{.05em}
DIST~\cite{dist} & 23.40 & 71.99\\
DIST++~\cite{dist}  & 17.52 & 72.65\\
Ours & \textbf{5.36} & \textbf{89.05}\\
\specialrule{.1em}{.05em}{.05em}
\end{tabular}
\vspace{-2mm}
\caption{\textbf{Image+LiDAR Completion on NorthAmerica.}} 
\label{tab:im-lidar}
\vspace{-6mm}
\end{table}

\vspace{-4mm}
\paragraph{Fine-tuning on sparse NorthAmerica dataset:} 
\sm{These
improvements demonstrated that the shape regularization provided by the discriminator makes the encoder and decoder more robust for real-world shape reconstruction. 
Encouraged by these results, we explored whether we could \shivam{further} improve performance by fine-tuning our model's encoder on sparse real-world data.
Specifically, we supervise the encoder to generate a latent code such that the decoder-predicted SDF values for the input observation points \wacvshivam{(sparse on-surface LiDAR points)} are 0.}
We also ensure that the shape latent-code generated by the encoder ($g_{\phi}(\mathbf{o}) = \mathbf{z}$) 
remains close to the shape latent-code generated by the pre-trained encoder without any fine-tuning ($\mathbf{z}_\text{iter-0}$). This serves as a regularization loss $(\cL^\text{z}_\text{finetune} = \|\mathbf{z} - \mathbf{z}_\text{iter-0}\|)$, preventing the encoder from generating an out-of-distribution latent-code. \shivam{We also regularize the reconstructed shape using
 discriminator-guided
  $\cL^\text{dis} (-\log D_{\psi}(\mathbf{X}, f_{\theta}(\mathbf{X}, \mathbf{z})))$ regularization term.}

Here we demonstrate shape completion results (``Ours$_{\text{finetune}}$'') obtained by fine-tuning \sm{our model's encoder} on sparse NorthAmerica dataset\footnote{\shivam{For fine-tuning/ training on NorthAmerica, we generate {3100} instances, which are completely different from NorthAmerica test samples.}}.
From Tab.~\ref{tab:quan-na}, we see that fine-tuning 
reduces the ACD by 18\% (Ours$_\text{finetune}$ vs Ours$_\text{no-finetune}$). \shivam{This showcase the effectiveness of our learned priors which allow us to fine-tune our model even using sparse on-surface LiDAR points as weak supervision. On the other hand}, prior (auto-decoder based) 
optimization methods like DeepSDF \cite{park2019deepsdf} and DIST \cite{dist} cannot be \sm{trained or fine-tuned on real data easily. 
This is because (1) real-world data lacks accurate dense supervision, and (2) DeepSDF and DIST do not} have a pre-trained representation of the latent code for real-world objects. 

\begin{table}[t]
\centering
\scalebox{0.85}{
\begin{tabular}{ccc|ccc|cc}
\specialrule{.2em}{.1em}{.1em}
\multicolumn{3}{c|}{Syn. Training Stage} & \multicolumn{3}{c|}{Test-time Opt. Stage} & ACD $\downarrow$ & Recall $\uparrow$\\
Dec. & Enc. & Disc. & Dec. & Enc. & Disc. & (mm) & (\%) \\
\specialrule{.1em}{.05em}{.05em}
\checkmark & & & \checkmark & & & 8.34 & 84.71 \\
\checkmark & \checkmark & & \checkmark & \checkmark & & 7.30 & 86.21 \\
\checkmark & \checkmark & \checkmark & \checkmark & \checkmark & & {6.96} & {86.55} \\
\checkmark & \checkmark & \checkmark & \checkmark & & \checkmark & 8.26 & 84.71 \\
\checkmark & \checkmark & \checkmark & \checkmark & \checkmark & \checkmark & 7.02 & 86.48 \\
\specialrule{.1em}{.05em}{.05em}
\end{tabular}}
\vspace{-2mm}
\caption{\textbf{Ablation study on encoder and discriminator.}}
\label{tab:enc-dec-ablation-training}
\vspace{-6mm}
\end{table}

\vspace{-2mm}
\paragraph{Qualitative Results:} Fig~\ref{fig:qual-NorthAmerica} compares Ours$_\text{finetune}$ reconstruction results with the prior works on NorthAmerica dataset. 
GRNet generates non-watertight shapes and fails to recover fine details. 
ONet produces overly-smooth shapes at times and doesn't have high-fidelity to the observations.  DeepSDF maintains high-quality local details for visible regions, \sm{but doesn't predict the correct shape}. 
Our approach produces results that are both visually appealing and have high-fidelity with the observations. 
Fig~\ref{fig:qual-KITTI} compares Ours$_\text{no-finetune}$ 
results with the prior works on KITTI, \sm{where we observe similar trends}. 
However, when input is very sparse, the reconstruction fails to match with the GT shape. 

\subsection{Image-based 3D Reconstruction} 
\label{sec:exp_image}
Our proposed pipeline can be easily extended to reconstruct shapes using other sensor modalities. 
Here, instead of encoding point-clouds, we demonstrate 3D reconstruction results by encoding a single-view image observation on the NorthAmerica dataset. 
Since the synthetic dataset we used does not have realistic texture maps for the corresponding 3D CAD models, we directly train the image encoder and discriminator on the real-world dataset. 
The training procedure is \sm{the same as}
stage 2 of the proposed curriculum learning technique \sm{(Sec.~\ref{sec:curriculum-learning}), but using an image encoder and using the real world GT shapes (same as those used for lidar-based fine-tuning) as supervision.} \shivam{Please check supp. for detailed training procedure.}
We compare our image-reconstruction results with state-of-the-art DIST\cite{dist} approach on the NorthAmerica dataset in Tab.~\ref{tab:im}. 
Compared to DIST, \shivam{our image-based}
approach produces significantly better results (85\% reduction in ACD), especially \sm{for occluded regions.} 
\sm{Moreover,} our image-only method \wacvshivam{(Tab.~\ref{tab:im} last-row)} is \sm{competitive with or better than all LiDAR completion baselines \wacvshivam{(Tab.~\ref{tab:quan-na} baselines)}.}

\subsection{Image + LiDAR Shape Completion} 
\label{sec:exp_lidar_image}
\sm{Our approach is amenable to combining sensor modalities for shape reconstruction without any re-training.}
We simply use the two pre-trained encoder modules (point-cloud encoder and image-encoder) and generate two 
shape latent-codes, $\mathbf{z}_\text{img}^\text{init}$ and $\mathbf{z}_\text{LiDAR}^\text{init}$. 
Then, inspired by 
GAN inversion \cite{gu2020image} and photometric stereo \cite{chen2018ps}, we 
fuse the two latent codes at the decoder features level.
Please refer to supp. for more details.
As shown in the Tab.~\ref{tab:im-lidar}, our proposed image + LiDAR shape completion technique performs significantly better than DIST and DIST++.
 Moreover, using both LiDAR and image observations reduces the ACD error by 10\% and 40\% compared to LiDAR and image only 
models.

\subsection{Analysis}
\label{sec:exp_ablation}

\begin{figure}
\centering
\includegraphics[width=0.80\linewidth]{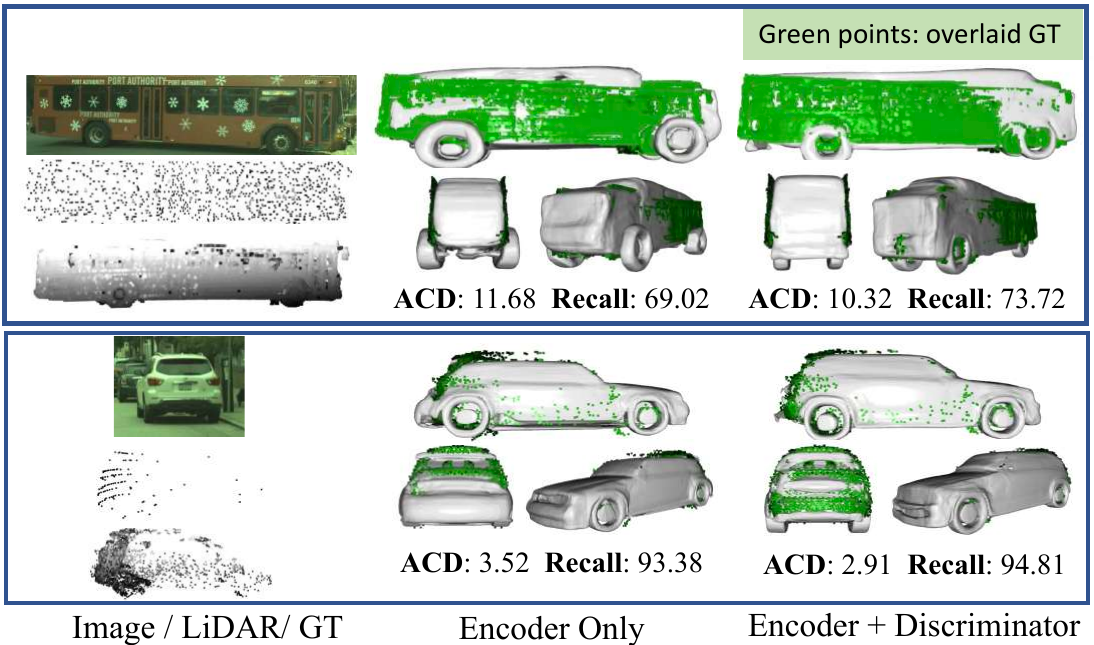}
\vspace{-2mm}
\caption{\textbf{Analyzing the effectiveness of discriminator} The discriminator improves the fidelity of the reconstructed shapes, even without any real-world training.
\siva{Let's only show one and show fewer views, make the figure simpler and easier to read}}
\label{fig:qual-discriminator}
\vspace{-6mm}
\end{figure}

\paragraph{Ablation study on Encoder and Discriminator:} 
\label{sec:enc-dis-ablation} 
We analyse the significance of the encoder (Enc.) and the discriminator (Disc.) in Tab.~\ref{tab:enc-dec-ablation-training} and in Fig.~\ref{fig:qual-discriminator}. To showcase our ``generalizability'', no real-world \sm{fine-tuning was performed.}
Adding both Enc. \sm{(row 1 vs. row 2)} and Disc. \sm{(row 2 vs. row 3)} to the reconstruction pipeline helps. \wacvshivam{Comparing row 3 with row 5 shows that adopting Disc. during optimization slightly affects quantitative performance, possibly due to noise in real-world aggregated LiDAR GT.}
\wacvshivam{However,} 
row 1 of Fig.~\ref{fig:qual-discriminator} \sm{demonstrates} that Disc. doesn't hinder generalization to unseen shapes\wacvshivam{, rather enhances the visual fidelity of the reconstructed shapes} (eg: model was trained on ShapeNet cars and tested on NorthAmerica bus category).

%% file: tex_files/conclusion.tex
\section{Conclusion}
In this paper, we present a simple yet effective solution for 3D object reconstruction in the wild. 
Unlike previous approaches that suffer from sparse, partial and potentially noisy observations, our method can generate accurate, \sm{high-quality} 3D shapes from real-world sensory data. 
The key to success is to fix the inherent 
limitations of existing methods: 
\wacvshivam{lack of robust shape priors and lack of proper regularization.}
To address these limitations, we highlight the importance of (i) \wacvshivam{deep encoder as a robust initializer of the shape latent-code;} (ii) \wacvshivam{regularized} test-time optimization of the latent-code; (iii) learned \wacvshivam{high-dimensional} shape regularization. 
\wacvshivam{Our proposed curriculum training strategy allows us to effectively learn disentangled shape priors. We significantly improve the shape reconstruction quality and achieve state-of-the-art performance on two real-world datasets.}

%% file: supp/abstract.tex

\section{Abstract} 
\label{sec:exp}

In this supplementary material, we first provide additional experimental details, which include dataset preparation details, implementation details and baselines' descriptions. (Sec.~\ref{sec:exp_details}). Then, we provide additional experimental results in Sec.~\ref{sec:exp_results} covering both quantitative and qualitative analysis for the three shape reconstruction tasks, we defined in the paper: LiDAR-based Shape Completion (Sec.~\ref{sec:exp_results_1}), Monocular Image Reconstruction (Sec.~\ref{sec:exp_results_2}) and Image + LiDAR Shape Completion (Sec.~\ref{sec:exp_results_3}). 

%% file: supp/experimental_details.tex

\section{Additional Experimental Details} 
\label{sec:exp_details}

In this section, we first provide additional data preparation 
details for all three datasets \sm{(ShapeNet, NorthAmerica, KITTI)}. Next, we describe the implementation and training details for our model as well as the baselines for the in-the-wild LiDAR shape completion task.

\subsection{Dataset Details}

\paragraph{ShapeNet:} We utilize {2364} shapenet meshes for training our shape reconstruction pipeline.  
To generate the ground-truth SDF volumes for supervising the training stages (stage 1 and stage 2), we first normalize each mesh within a unit sphere and then randomly sample {16384} points from the sphere. For each sampled point, GT SDF value is generated by measuring its distance from the mesh surface along the GT surface normal. We also render each mesh at 5 different viewpoints to generate the sparse input point clouds required to train the LiDAR encoder in Stage 2. The viewing transformations used for rendering are randomly selected \sm{poses} from the NorthAmerica dataset. The rendered depth maps are also sampled using the NorthAmerica's LiDAR sensor's azimuth and elevation resolutions. These sampled depth maps are then unprojected to generate the sparse point clouds. We used Blender's cycles engine to render the meshes.

\paragraph{NorthAmerica:} NorthAmerica dataset has long trajectories of sparse point clouds captured by a 64-line spinning 10Hz LiDAR sensor and RGB images captured by a wide-angle perspective camera, with both the sensors synchronized in time. Centimeter level localization and manually annotated object 3D bounding boxes are also created through an off-line process. These provide accurate poses and regions-of-interest to aggregate individual LiDAR sweeps and produce dense scans that serve as our GT {for evaluation}. Specifically, we exploit these manually annotated ground-truth detection labels to extract object-specific data from these raw sensor data. Despite using ground-truth bounding boxes, the extracted object data is still noisy (\ie contains road points, non-car movable objects, etc). To reduce the noise in the extracted point cloud, we filter out the non-car points using a trained LiDAR segmentation model \cite{lidar_segmentation}. \siva{cite the model, should be chris's paper} The ground-truth shape is generated by aggregating multi-frame (maximum of 60 frames, captured at a rate of 10Hz) object LiDAR points in the object-coordinate space. If the object LiDAR points are less than 100 points for all frames, we discard the object. We further symmetrize the aggregated point cloud along the lateral dimension to create a more complete GT shape. For dynamic objects, we additionally perform color-based ICP \cite{color-icp} (using LiDAR intensity values) to better register the multi-sweep data. \siva{cite the color ICP  paper - check open3d docs reference }

\paragraph{KITTI:} We use KITTI's ground-truth bounding boxes to aggregate the multi-sweep object data in the object-coordinate space. Since KITTI bounding boxes are too tight, \sm{which causes parts of the object to not be included during aggregation}, we expand the bounding boxes by $10\%$ along each dimension. The aggregated objects in KITTI dataset are noisier (contains interior points, flying 3D points) compared to the NorthAmerica dataset. To reduce the noise, we first perform spherical filtering by filtering out all the points, which lie at a distance greater than a specified threshold from the center of the object. Next, we apply statistical (neighbor density based) outlier removal to filter out the isolated points. Like NorthAmerica dataset, we symmetrize the aggregated point cloud along the lateral dimension. 

\subsection{Implementation Details}
\label{sec:lidar-completion-training}
\textbf{Curriculum Training and Inference procedure:} 1) \textbf{Training Stage 1:} Our shape reconstruction pipeline consists of encoder, decoder and discriminator module. As described in the main paper, we use a {2}-stage curriculum learning strategy to train our model on the synthetic Shapenet dataset. In the first stage, similar to DeepSDF\cite{park2019deepsdf}, we train an auto-decoder framework to jointly optimize the ({256}-dimensional) shape latent-code and the decoder weights. The loss function for stage 1 training is: $\cL^\text{dec} \;+\; \cL^\text{reg}$, where $\cL^\text{dec}$ loss term is supervised by the ground-truth SDF volumes (with {16384} randomly sampled points). We train the auto-decoder framework for 1000 epochs in Stage 1. 
2) \textbf{Training Stage 2:} Next, we train the encoder and the discriminator (keeping the decoder fixed) for 280 epochs in stage 2. The input to the LiDAR encoder is a partial point-cloud with \emph{k} 3D points ($\emph{k} \leq 1024$). 
The encoder then outputs a 256-dimensional latent-code. Given this latent-code and a grid of 3D points, the decoder predicts the corresponding SDF volume. The discriminator then classifies the predicted SDF volume as real or fake. For training the discriminator, we first generate real and fake SDF volumes. Both real and fake SDF volumes use the same set of 3D points. These 3D points are randomly sampled from the whole 3D space and are augmented with the \emph{k} input points, leading to a total of {4096} points. For the real samples, the SDF values corresponding to the 3D points are generated using stage 1's pre-trained latent-codes, while for the fake SDF volumes we use the encoder predicted latent-code. The loss function for stage 2 training is: $\cL^\text{dec} + \cL^\text{z} + \cL^\text{gan}$, where $\cL^\text{dec}$ is same as stage 1's $\cL^\text{dec}$ loss term. $\cL^\text{z}$ loss term is supervised by the trained latent-codes of stage 1. $\cL^\text{gan}$ loss term which uses the real and fake SDF volumes, guides the optimization of discriminator weights. For both the training stages, we weight the $\cL_\text{dec}$, $\cL_\text{gan}$ and $\cL_\text{reg}$(or $\cL_\text{z}$) terms in the ratio of 2:1:1.
3) \textbf{Inference:} Finally, at test-time, given in-the-wild real-world observations, we utilize the trained encoder module to predict an initial latent-code. This latent code is then optimized for 800 iterations using the inference procedure defined in the paper Sec. 4.2. We use first-order gradient optimizer \cite{kingma2014method} for both training and inference. During inference also, we weight $E_\mathrm{data}$, $E_\mathrm{reg}$ and $E_\mathrm{dis}$ in the ratio 2:1:1.

\paragraph{Architecture Details:} 1) \textbf{Decoder}: 
\shivam{The decoder architecture is exactly same as DeepSDF \cite{park2019deepsdf}.}\siva{I dont understand this, isn't stage 1 training the decoder}
2) \textbf{LiDAR Encoder}: We exploit the stacked PointNet network proposed by Yuan \etal. \cite{yuan2018pcn} as our encoder. More specifically, the first PointNet \cite{qi2017pointnet} block has 2 layers, with 128 and 256 units respectively. The second PointNet block has 2 layers with 512 and 1024 units. The second block takes as input both the individual point features and the global max-pooled features. The stacked PointNet encoder is followed by a final fully-connected layer (and BatchNormalization) with 256 units, with tanh as the final activation. 
3) \textbf{Discriminator}: The architecture of our discriminator is akin to PointNet.
It consists of 5 fully-connected layers (with 64, 64, 64, 128, 1024 units) for computing per-point features, and a final global max-pooling layer. We separately compute batch-norm \cite{batchnorm} mean and variance statistics for the real and the fake samples.

\subsection{Baselines}

\paragraph{DeepSDF \cite{park2019deepsdf}:} We follow the exact same implementation and point sampling guidelines as \cite{park2019deepsdf} to train DeepSDF on ShapeNet. In particular, we exploit 16384 SDF samples within a unit sphere enclosing the object for optimization.  The shape latent code has a dimensionality of 256. The decoder is an 8 layer fully connected MLP, with ReLU as intermediate activations and tanh as the activation of the last layer. Please refer to Park \etal. \cite{park2019deepsdf} for more architectural details. As for the real-world dataset, we optimize the SDF loss using the on-surface LiDAR points and the randomly sampled off-surface points. We use a maximum of 1024 on-surface points. The off-surface points are randomly sampled within a truncated distance of $\pm$ 0.02, along the rays joining the object center and the surface LiDAR points.

\paragraph{ONet \cite{ONet}:} We follow the same implementation guidelines as mentioned by the authors, except that we use 16384 occupancy samples to supervise the reconstruction of each shape during training.

\paragraph{GRNet \cite{xie2020grnet}:} We use the released codebase and the ShapeNet pre-trained weights to reconstruct shapes on KITTI and NorthAmerica datasets.

\paragraph{DIST/DIST++: \cite{dist}} For LiDAR completion task, DIST requires ground-truth depth maps for supervising their auto-decoder optimization framework.
The ground-truth sparse depth map was generated by projecting the single-sweep LiDAR points onto the camera image using ground-truth poses. For DIST++, we augment the original DIST optimization strategy with the SDF loss on off-surface points.

\paragraph{SAMP \cite{Engelmann_2017}:} We first construct a PCA embedding based on 103 CAD models purchased from TurboSquid \cite{turbosquid}. We use the purchased models instead of ShapeNet since a good PCA embedding requires shapes to be in a proper metric space and to be watertight to compute proper volumetric SDFs. Most of the purchased CAD models are passenger vehicles such as sedans and mini vans.
We compute volumetric SDFs for each vehicle in metric space, where the output volume has dimensions $(300 \times 100 \times 100)$. The resolution of each voxel is 0.025 meters. We set the embedding dimension to be 25, and optimize the shape embedding as well as a scaling factor on the SDF to handle larger shapes. We use Adam with learning rate 0.2.  
The loss function includes a smooth L1 data term, an L2 regularization term, and a scale factor regularization term. The weights of these terms are 1, 0.05, and 0.01, respectively.

%% file: supp/experimental_results.tex

\section{Additional Experimental Results} 
\label{sec:exp_results}

\subsection{LiDAR-based Shape Completion}
\label{sec:exp_results_1}

In this section, we compare all the LiDAR-based shape completion approaches both quantitatively (Cumulative ACD and Recall comparison) and qualitatively. 

\begin{figure*}[htb]
\centering
\begin{subfigure}[c]{0.45\textwidth}
\includegraphics[width=\linewidth,trim={0mm 0mm 0mm 0mm},clip]{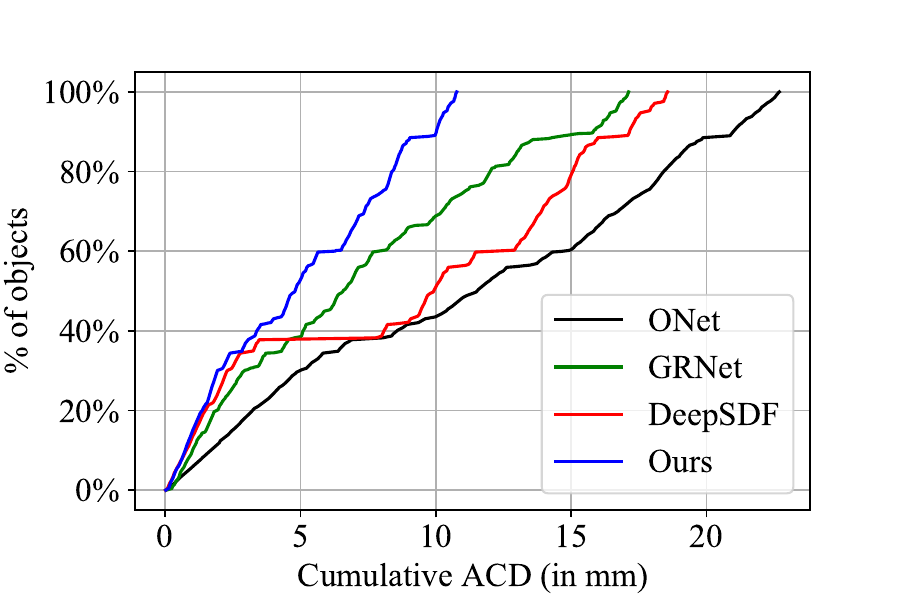}
\caption{KITTI Dataset}
\label{fig:cumulative-acd-kitti}
\end{subfigure}
\begin{subfigure}[c]{0.45\textwidth}
\includegraphics[width=\linewidth,trim={0mm 0mm 0mm 0mm},clip]{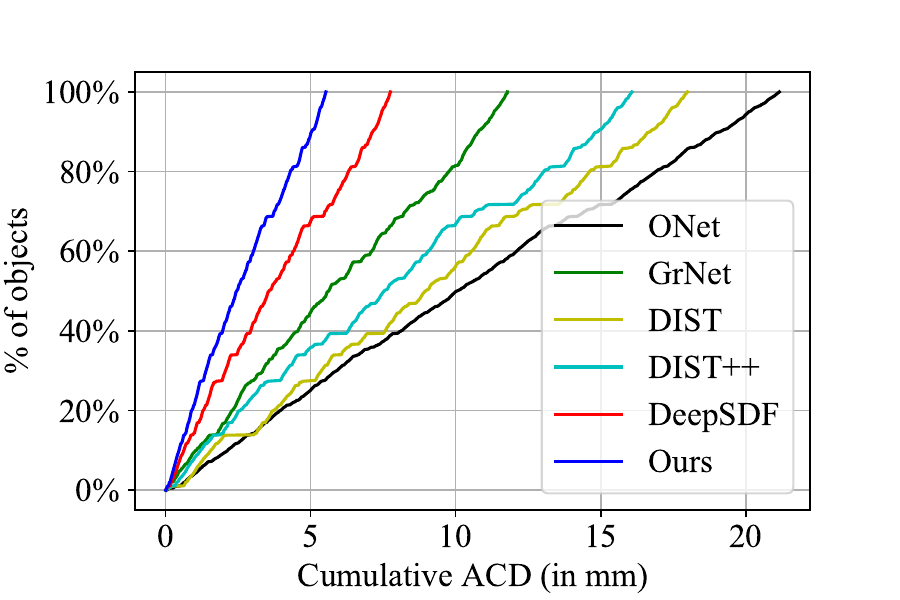}
\caption{NorthAmerica Dataset}
\label{fig:cumulative-acd-NorthAmerica}
\end{subfigure}
\caption{Cumulative Asymmetric Chamfer Distance for LiDAR Completion}
\end{figure*}

\paragraph{Cumulative ACD Analysis:} Fig.~\ref{fig:cumulative-acd-kitti} and Fig.~\ref{fig:cumulative-acd-NorthAmerica} compares the cumulative ACD among various 3D shape completion approaches on KITTI and NorthAmerica datasets respectively. Our approach consistently outperforms all prior work. Compared to encoder-based initialization approaches, DeepSDF suffers from a sudden jump in the cumulative error on the KITTI dataset. We conjecture this is because some of the aggregated ground-truth objects are in fact noisy.

\begin{figure*}[htb]
\centering
\begin{subfigure}[c]{0.45\textwidth}
\includegraphics[width=\linewidth,trim={0mm 0mm 0mm 0mm},clip]{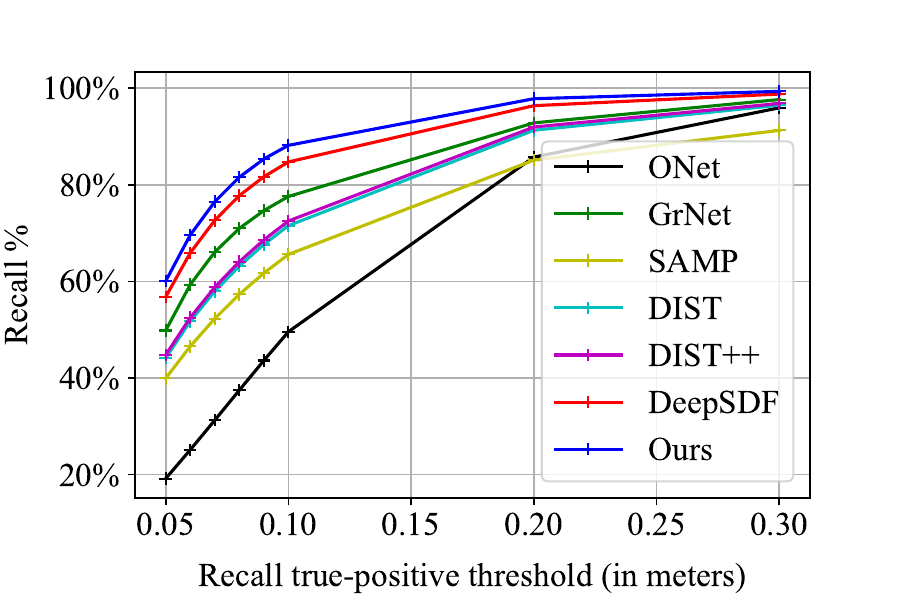}
\caption{True-positive threshold vs Recall}
\label{fig:recall-NorthAmerica}
\end{subfigure}
\begin{subfigure}[c]{0.45\textwidth}
\includegraphics[width=\linewidth,trim={0mm 0mm 0mm 0mm},clip]{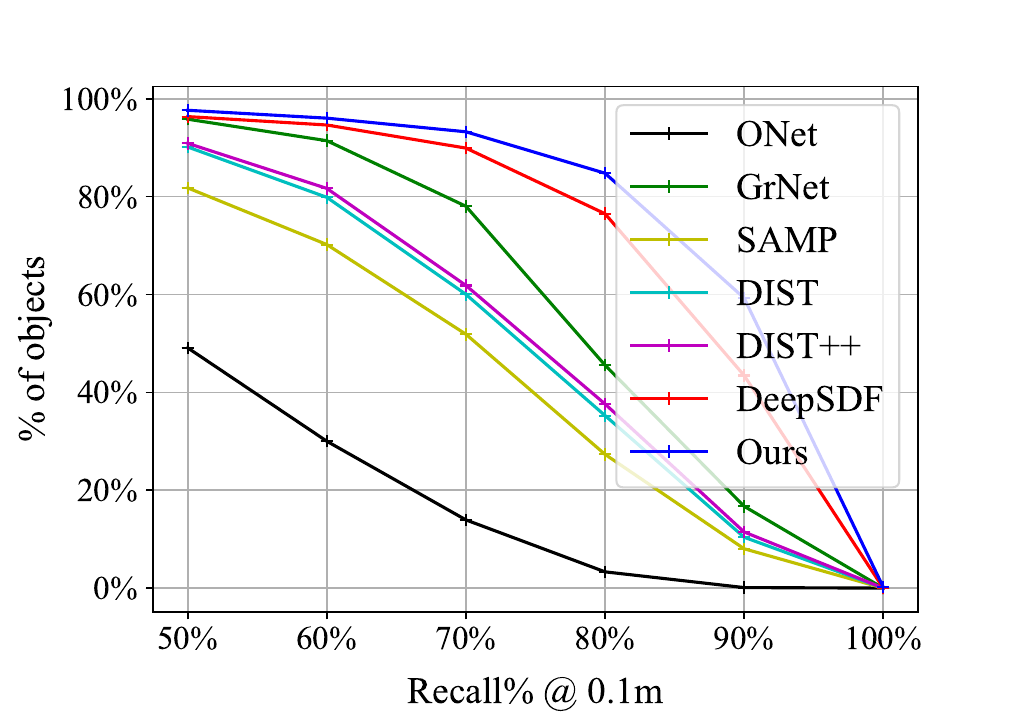}
\caption{Cumulative Recall}
\label{fig:recall-object-NorthAmerica}
\end{subfigure}
\caption{Recall Analysis for LiDAR Completion on NorthAmerica Dataset}
\siva{not sure if possible but can we remove the legend for part b and say in caption legend is shareed across the figure, currently the lines are covered and its hard to see them.}
\end{figure*}

\paragraph{Recall Analysis on NorthAmerica:} Fig.~\ref{fig:recall-NorthAmerica} showcase the variation of recall as a function of true-positive distance threshold. Fig.~\ref{fig:recall-object-NorthAmerica} analyzes the percentage of objects with recall greater than or equal to a certain value, at a distance threshold of 0.1m. Approximately, 98\% of our reconstructed objects have recall $>=$ 50\% and ~87\% of our reconstructed objects have recall $>=$ 80\%, comparing to 96\% and 76\% of DeepSDF's object respectively.

\begin{figure*}[htb]
\centering
\vspace{5mm}
\includegraphics[width=0.95\linewidth]{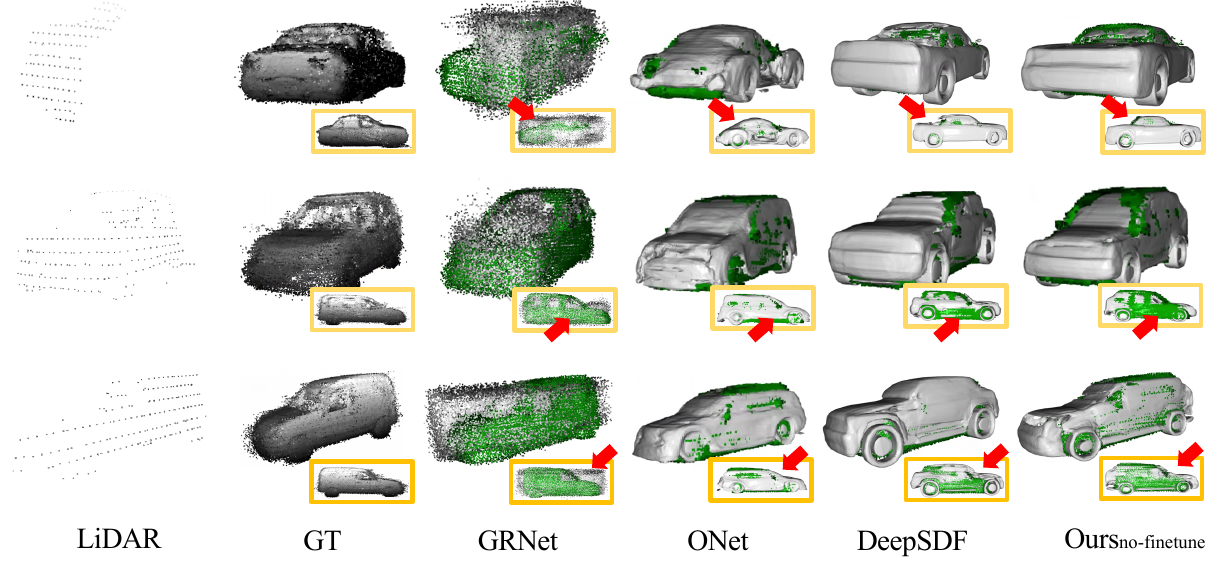}
\vspace{-1mm}
\caption{\textbf{LiDAR Completion on KITTI:} In the KITTI dataset, LiDAR point clouds are much noisy and sparse, and the vehicle bounding boxes needed to normalize the input data are not accurate either. Because of this all KITTI reconstructions have more artifacts compared to NorthAmerica reconstructions. Compared to all the other approaches, our shapes have much less artifacts and maintain finer details. }
\label{fig:lidar-comparison-KITTI}
\vspace{-5mm}
\end{figure*}

\begin{figure*}[htb]
\centering
\includegraphics[width=0.95\linewidth]{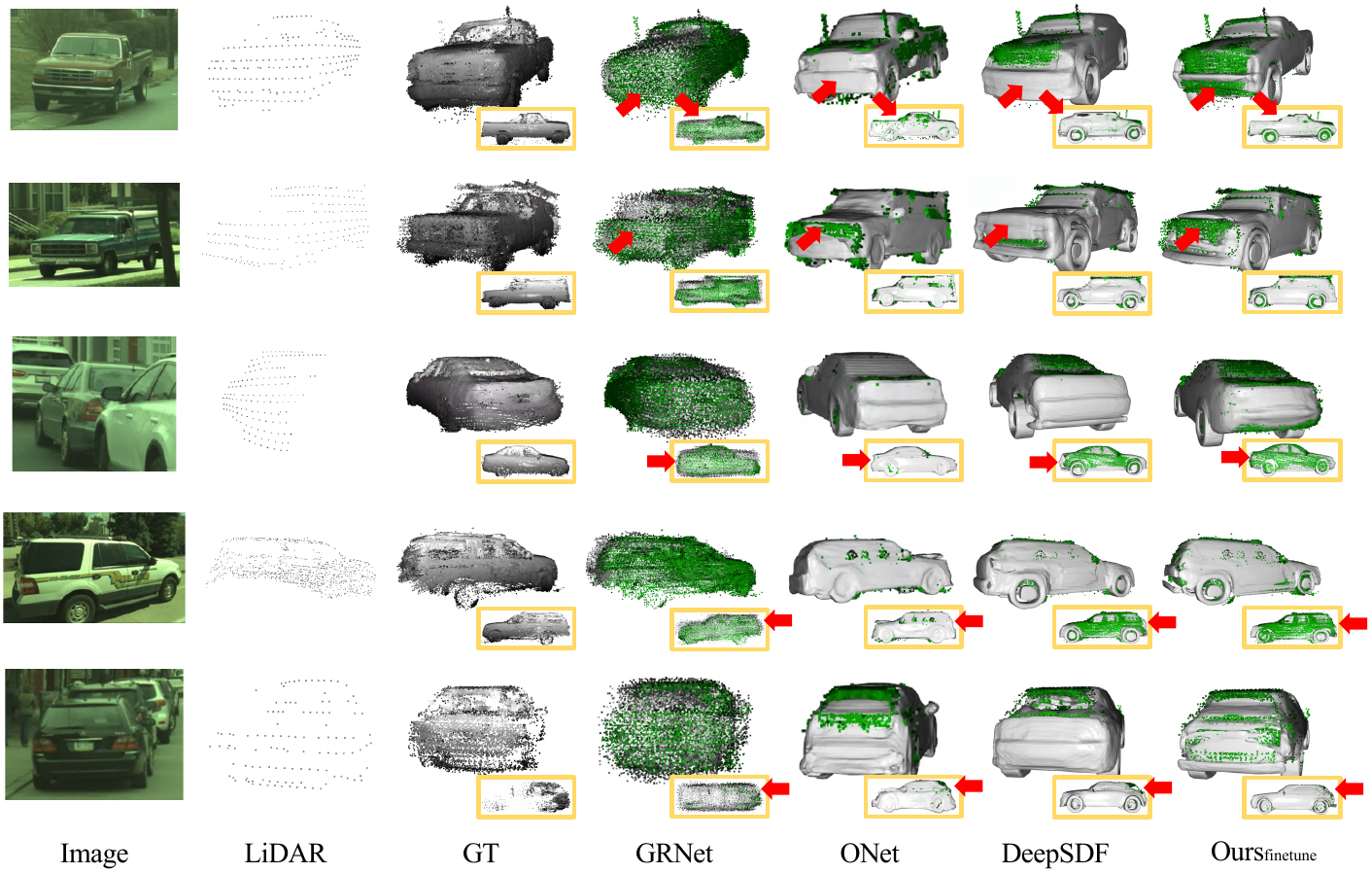}
\vspace{-1mm}
\caption{\textbf{LiDAR Completion on NorthAmerica:} Compared to prior works, our approach (1) maintains fine details, (2) register well with the input and the GT and, (3) works well with sparse and occluded observations. Please zoom in for enhanced visual comparison.}
\label{fig:lidar-comparison-NorthAmerica}
\vspace{-5mm}
\end{figure*}

\begin{figure*}[htb]
\centering
\vspace{5mm}
\includegraphics[width=0.70\linewidth]{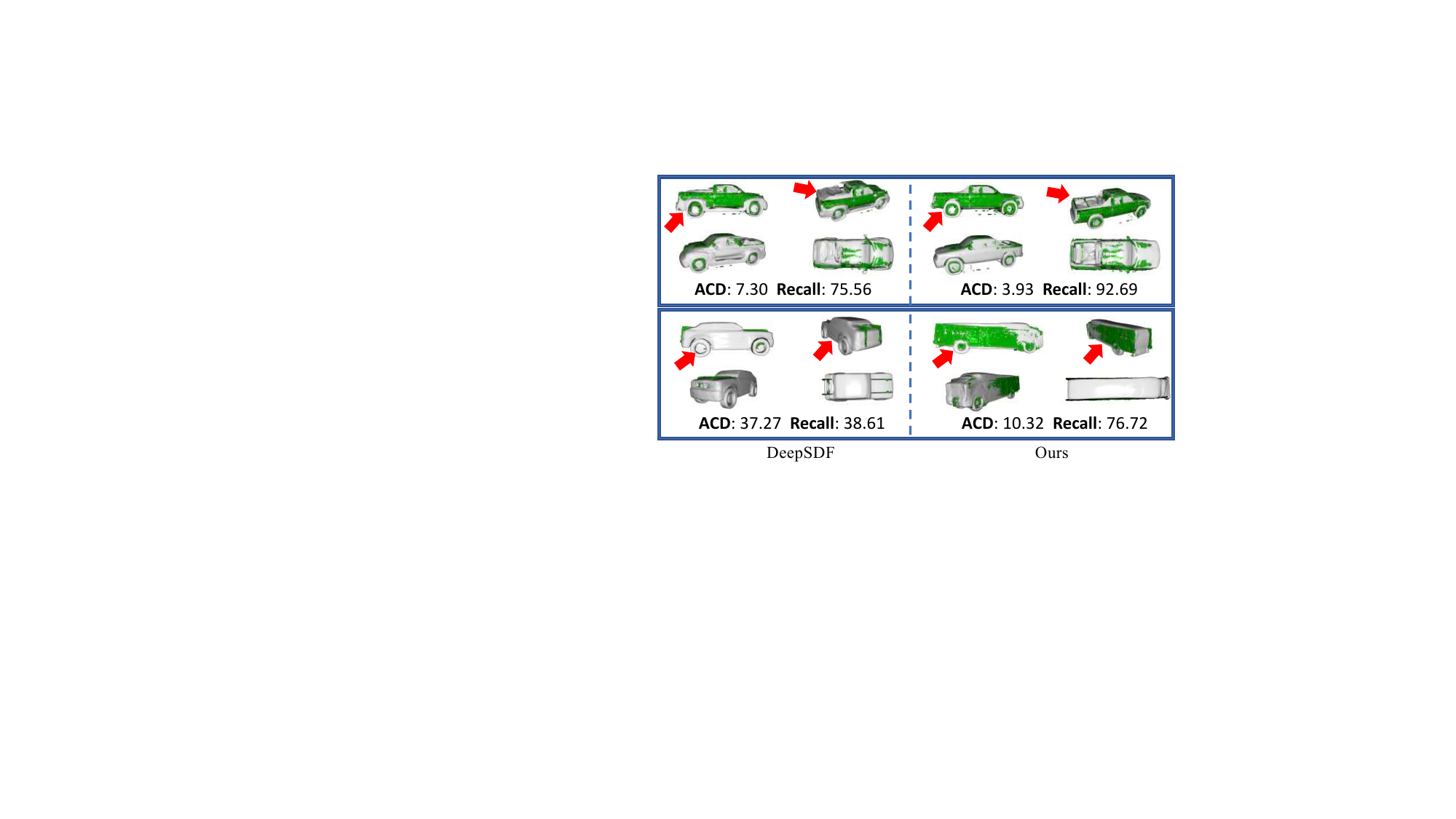}
\caption{\textbf{In-the-wild LiDAR Completion Comparison (Ours$_\text{no-finetune}$ vs DeepSDF) on NorthAmerica.} Our approach generates much cleaner shapes and maintains the fine details even without any real-world fine-tuning (eg: compare the cavity of the reconstructed pickup cars in Row 1). Our proposed framework is also generalizable to unseen vehicles (eg: check the unseen NorthAmerica dataset bus in Row 2).}
\label{fig:in-the-wild-lidar-comparison-NA}
\vspace{-5mm}
\end{figure*}

\paragraph{Qualitative Analysis:} Fig.~\ref{fig:in-the-wild-lidar-comparison-NA} and Fig.~\ref{fig:lidar-comparison-KITTI} compares Ours$_\text{no-finetune}$ results with the prior works on NorthAmerica and KITTI dataset respectively. \siva{Modify the figure (currently it just says Ours, ambiguous) to change the ours to Ours\_no-finetune or Ours\_finetune. Also no-finetune is a bit too long of a subscript} \sduggal{Done. We used no-finetune in the paper, so probably better to keep the same.} No fine-tuning on real-world datasets was done for these visualizations. Green points depict the overlay of the GT point clouds. As shown in Fig.~\ref{fig:in-the-wild-lidar-comparison-NA}, our approach creates much cleaner shapes, maintaining both the global structure and the fine details of the GT shape better than the state-of-the-art DeepSDF\cite{park2019deepsdf} method. Fig.~\ref{fig:lidar-comparison-NorthAmerica} further compares Ours$_\text{finetune}$ results (obtained by fine-tuning the ShapeNet pre-trained model on NorthAmerica dataset) with the prior works. While GRNet fails in completing the full shape, ONet tends to predict over-smooth shapes. Both DeepSDF and our approach produce complete and high-quality shapes, yet our reconstructed meshes have more fine-grained, structured details.

\subsection{Monocular Image Shape Reconstruction}
\label{sec:exp_results_2}

In this section, we first describe the training procedure used by our single-image 3D reconstruction approach. Next, we showcase some qualitative comparisons between our image reconstruction method and the state-of-the-art DIST\cite{dist} approach.

\paragraph{Training procedure:} Similar to our LiDAR completion model, our image reconstruction pipeline consists of an image encoder, decoder and discriminator. As mentioned in the paper Sec. {6.2}, we directly train our image encoder and discriminator on the real-world NorthAmerica dataset (training set of {3100} instances). Prior to training the image encoder, we first train our LiDAR completion model (LiDAR encoder, decoder and discriminator) on the Shapenet dataset as mentioned in Sec.~\ref{sec:lidar-completion-training}. We then utilize this pre-trained model and perform inference using real-world LiDAR observations to generate a shape latent-code (predicted by the LiDAR encoder) and a SDF volume (predicted by the decoder for a randomly sampled grid of 3D points). The predicted shape latent-codes serve as the pseudo-GT for supervising the training of the image encoder, and the predicted SDF volumes serve as the real input samples for training the discriminator. In addition to the pseudo-GT, we have sparse on-surface LiDAR points (with SDF=0) which serve as the GT SDF data. \shivam{Rest, following the training procedure mentioned in paper Sec. {4.3}, we train the image encoder and discriminator for 360 epochs on NorthAmerica dataset training set.}

\begin{figure}[!htb]
\centering
\resizebox{0.6\textwidth}{!}{%
\includegraphics[width=0.90\linewidth]{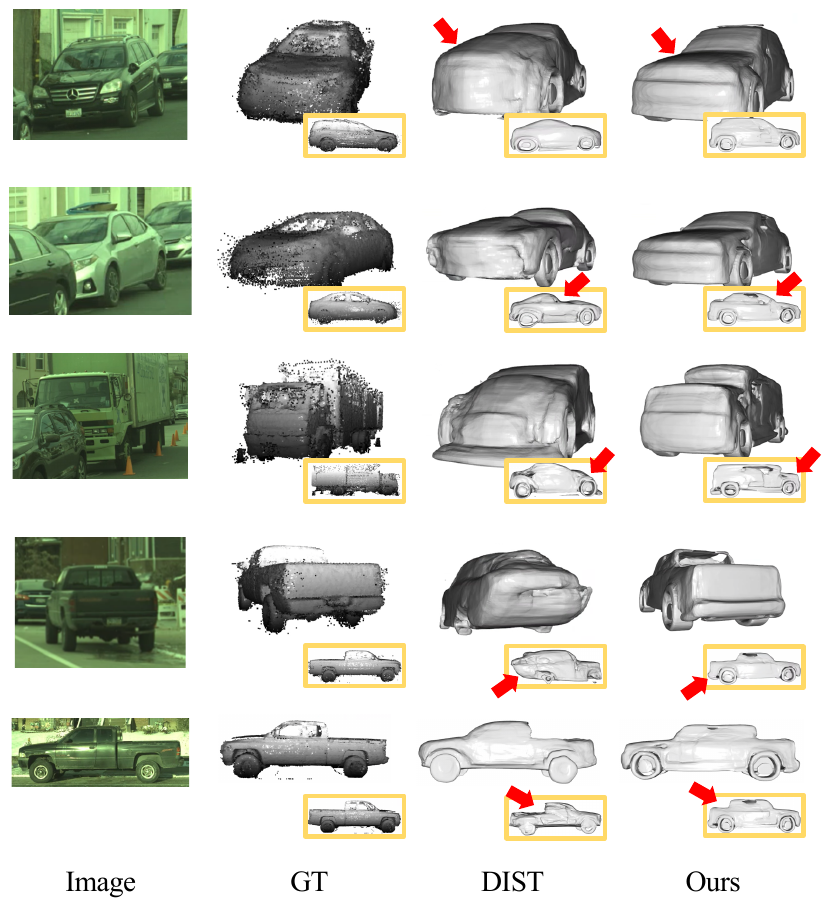}}
\vspace{-2mm}
\caption{\textbf{Image-based Reconstruction on NorthAmerica:} Thanks to the learned shape priors, we can easily train our model on real-world datasets using only sparse on-surface points as GT. Compared to DIST, our approach has much less artifacts and performs well even in the regions occluded in the input image. Check the gif visualizations in the attached video for multi-view comparson of the reconstructed shapes.}
\label{fig:image-comparison-NorthAmerica}
\vspace{-5mm}
\end{figure}

\paragraph{Qualitative Analysis:} Fig.~\ref{fig:image-comparison-NorthAmerica} compares image-based 3D reconstruction approaches on NorthAmerica dataset. 
For DIST\cite{dist}, we use two camera images to optimize the 3D shape and we optimize the photometric loss and the latent code regularization term during test-time optimization of the latent-code. Unlike DIST, we use only one single image to generate the shape latent code. Our results yet still have significantly higher fidelity than those of DIST. Since the image reconstruction model is trained using noisier real-world LiDAR data and the captured LiDAR is particularly more noisy near the transparent windows, we notice some holes near the vehicle windows in some of the reconstructed shapes. 
Please check the visual comparison in the attached video for multi-view (gif-based) visualization of the reconstructed shapes.

\subsection{Image + LiDAR Shape Completion} 
\label{sec:exp_results_3}
As mentioned in the paper Sec. {6.3}, our approach allows us to combine multiple sensor modalities for shape reconstruction without any re-training. We utilize a multi-code fusion technique to combine the image and LiDAR observations directly at test-time. In this section, we first provide a detailed description of the multi-code fusion and optimization technique, followed by its quantitative analysis. Finally, we showcase some qualitative comparisons and ablations for image + LiDAR 3D reconstruction.

\begin{figure*}[t]
\centering
\begin{subfigure}[c]{0.45\textwidth}
\includegraphics[width=\linewidth,trim={0mm 0mm 0mm 0mm},clip]{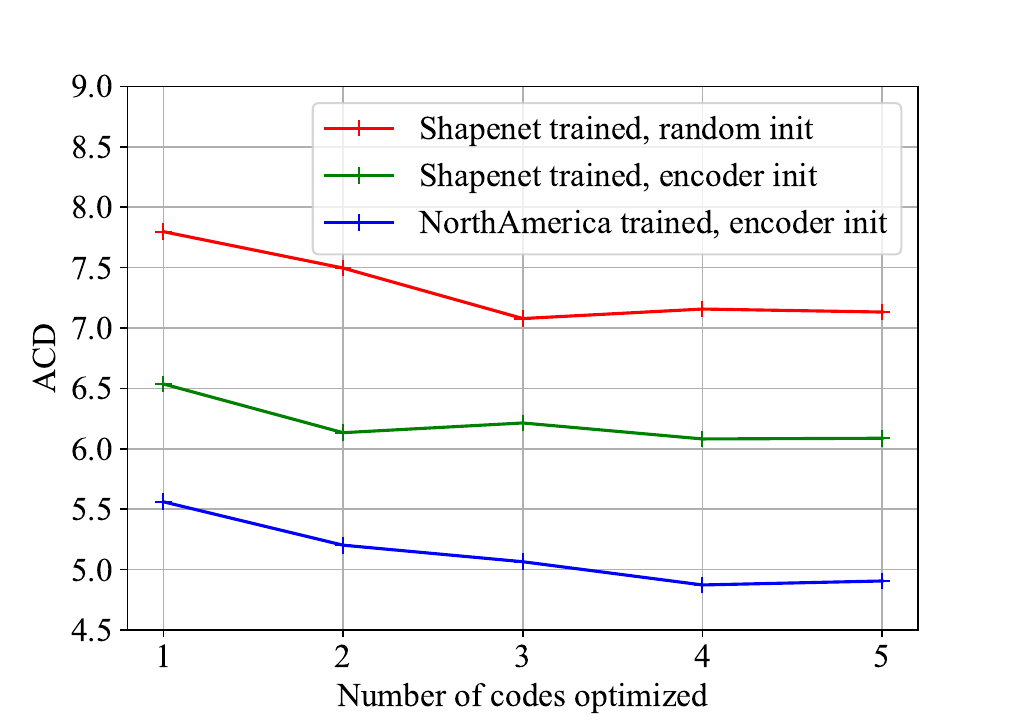}
\caption{ACD on NorthAmerica}
\label{fig:mcode-acd-NorthAmerica}
\end{subfigure}
\begin{subfigure}[c]{0.45\textwidth}
\includegraphics[width=\linewidth,trim={0mm 0mm 0mm 0mm},clip]{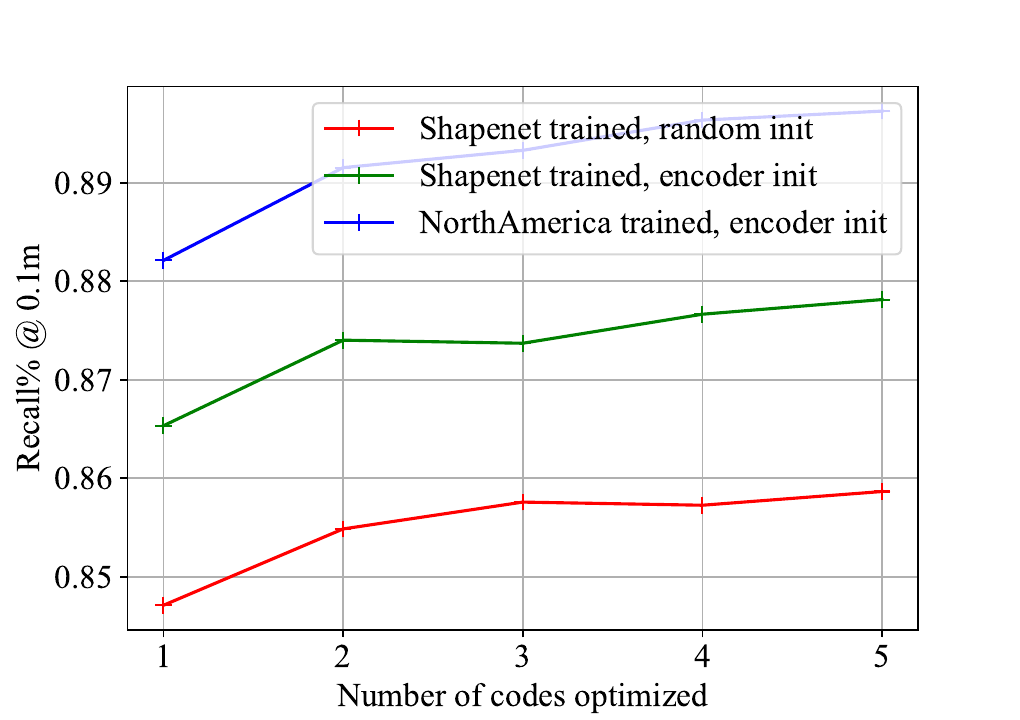}
\caption{Recall on NorthAmerica}
\label{fig:mcode-recall-NorthAmerica}
\end{subfigure}
\caption{Performance vs \# of codes in multi-code optimization}
\end{figure*}

\paragraph{Multi-code optimization:} Given Image and LiDAR observations, we simply use the two pre-trained encoder modules (point-cloud encoder and image-encoder) and generate two shape latent-codes, $\mathbf{z}_\text{img}^\text{init}$ and $\mathbf{z}_\text{LiDAR}^\text{init}$. Next, inspired by the latest effort on GAN inversion \cite{gu2020image} and photometric stereo \cite{chen2018ps}, we propose to fuse the two latent codes at the feature level. We first divide decoder module ($\mathbf{f_{\theta}}$) into two sub-networks $\mathbf{f}^{l}_{\theta_1}$ and $\mathbf{f}^{l}_{\theta_2}$, where $l$ indicates the layer that we split, $\theta_1$ and $\theta_2$ are parameters of the two sub-networks and $\mathbf{f}_{\theta} = \mathbf{f}^{l}_{\theta_2} \circ \mathbf{f}^{l}_{\theta_1}$. Then we pass both latent codes into the first sub-network, $\mathbf{f}^{l}_{\theta_1}$, and generate two corresponding feature maps. Finally the estimated feature maps are aggregated and passed into the second sub-network $\mathbf{f}^{l}_{\theta_2}$. Using this multi-code fusion procedure, the inference procedure becomes:
\begin{align}
\begin{split}
\mathbf{z}^\ast = &\argmin_{\mathbf{z}_\text{img}, \mathbf{z}_\text{LiDAR}} E_\mathrm{data}(\bo, \mathbf{f}^{l}_{\theta_2}(\mathbf{X}, \mathbf{z}_l)) + \lambda_\text{reg} E_\mathrm{reg}(\mathbf{z}_\text{img}) \\ 
&+ \lambda_\text{reg} E_\mathrm{reg}(\mathbf{z}_\text{LiDAR}) + \lambda_\text{dis}E_\mathrm{dis}(\mathbf{f}^{l}_{\theta_2}(\mathbf{X}, \mathbf{z}_l)),
\end{split}
\end{align}
where $\mathbf{z}_l = g(\mathbf{f}^{l}_{\theta_1}(\mathbf{x}, \mathbf{z}_\text{img}),~\mathbf{f}^{l}_{\theta_1}(\mathbf{x}, \mathbf{z}_\text{LiDAR}))$ is the aggregated feature and $g$ is the aggregation function. $\mathbf{z}_\text{LiDAR}$ and $\mathbf{z}_\text{img}$ are initialized using $\mathbf{z}_\text{LiDAR}^\text{init}$ and $\mathbf{z}_\text{img}^\text{init}$ respectively. $E_\mathrm{data}$, $E_\mathrm{reg}$ and $E_\mathrm{dis}$ are same as defined in paper Sec. {4.2}. Following \cite{chen2018ps}, we adopt {max-pooling} to fuse the features. The advantages of the proposed multi-fusion technique are three fold: (i) there are no extra parameters required for multi-sensor fusion; (ii) the aggregation function can be extended to take multiple features as input without modification; and (iii) max-pooling over the features allows each latent code to focus on the part it is confident in and distribute the burden accordingly.

\paragraph{Quantitative Analysis:} The proposed multi-code optimization technique can even be used with a single sensor observation. Given an observation (eg: an image or a LiDAR point cloud), we first generate an initial latent code. Then, we generate $n$ different latent codes by jittering the initial code using multiple normally-distributed noise vectors. We then jointly optimize these codes using our proposed multi-code optimization strategy, and fuse them in the same way we fused Image and LiDAR generated shape codes. The same multi-code optimization procedure can be performed for DeepSDF approach, where instead of one, we optimize $n$ different random latent-codes using the auto-decoder framework. Fig.~\ref{fig:mcode-acd-NorthAmerica} and Fig.~\ref{fig:mcode-recall-NorthAmerica} showcase the performance boost achieved by optimizing multiple codes, which is irrespective of the latent-code initialization and the training dataset used. We observe a boost of around 1.3-1.7\% on recall, and a decrease of around 8-10\% on ACD, when we optimized four latent codes instead of one, for all the experimental settings.

\begin{figure}[!htb]
\includegraphics[width=0.90\linewidth]{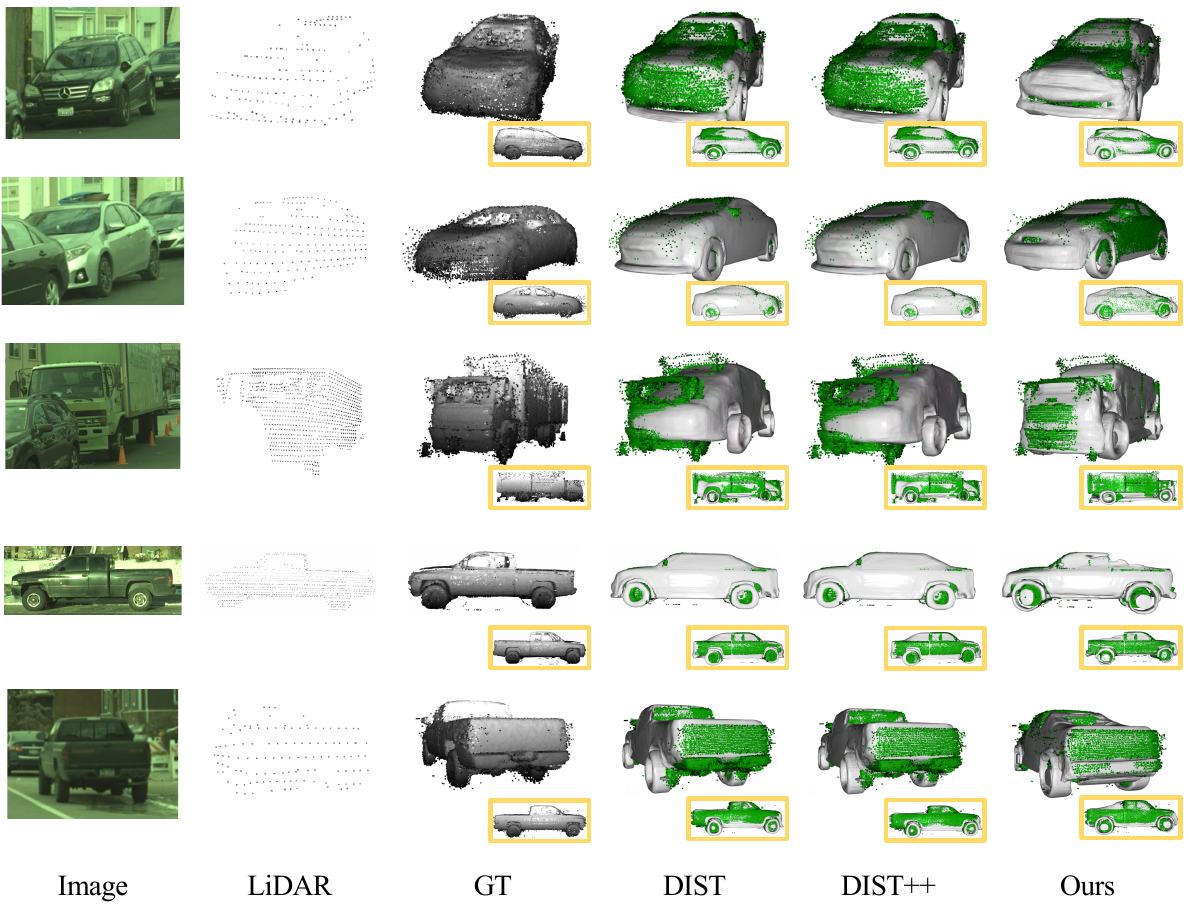}
\vspace{-2mm}
\caption{\textbf{Image+LiDAR Shape Completion on NorthAmerica}}
\label{fig:image-lidar-comparison-NorthAmerica}
\vspace{-5mm}
\end{figure}

\paragraph{Qualitative Comparison:} Fig.~\ref{fig:image-lidar-comparison-NorthAmerica} compares Image + LiDAR Shape Completion approaches on NorthAmerica. Adding LiDAR to image-only reconstruction pipelines helps all the approaches (DIST/ DIST++/ Ours) in generating shapes with higher fidelity. Thanks to the proposed neural initialization/ regularized optimization, our method's predicted shapes align well with GT shape and maintain much finer details.

\begin{figure*}[!htb]
\centering
\includegraphics[width=0.95\linewidth]{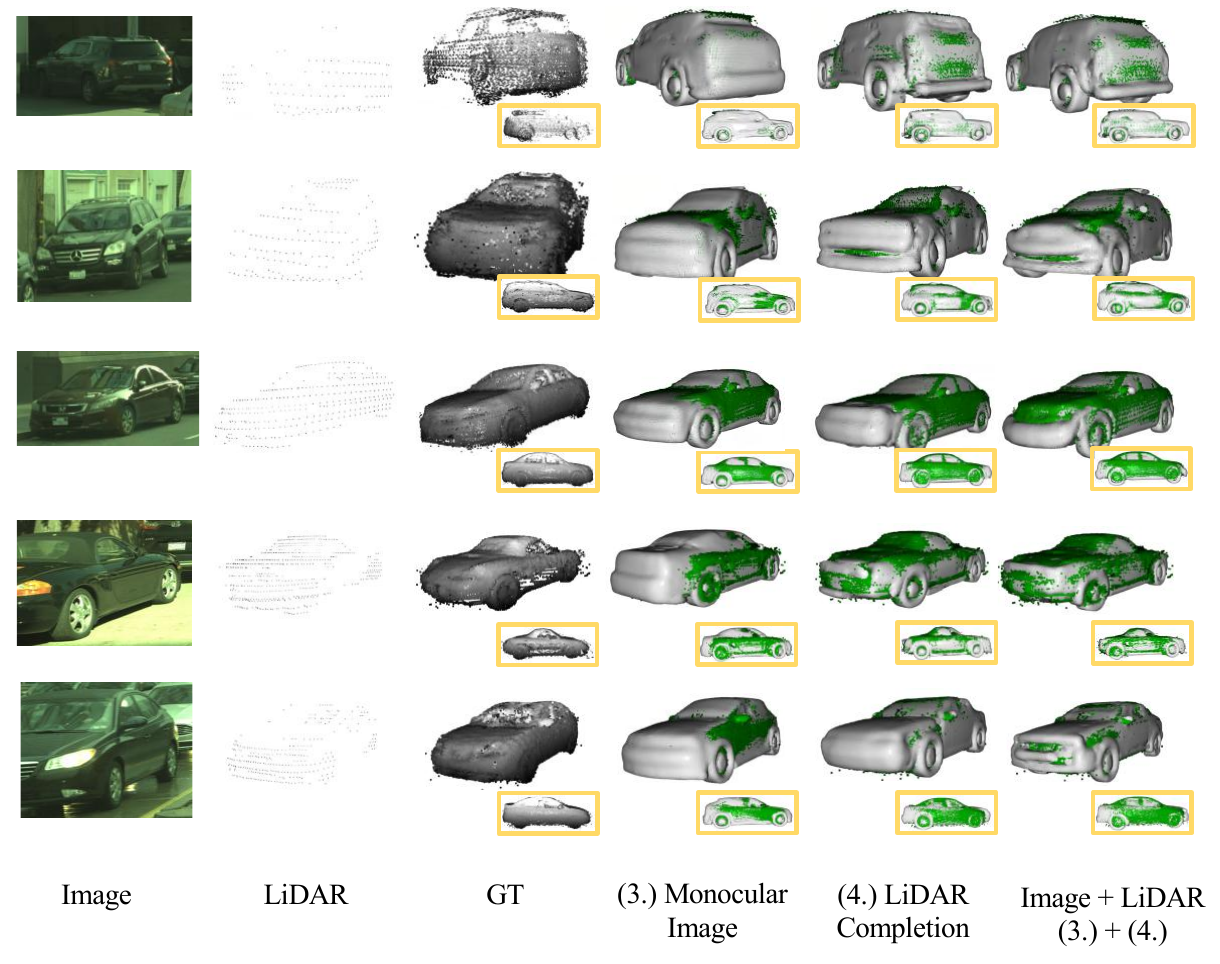}
\vspace{-2mm}
\caption{\textbf{Image+LiDAR ablation on NorthAmerica}}
\label{fig:image-lidar-ablation-NorthAmerica}
\vspace{-5mm}
\end{figure*}

\paragraph{Ablation Study on Image + LiDAR 3D Reconstruction on NorthAmerica:} Fig.~\ref{fig:image-lidar-ablation-NorthAmerica} visually compares the shapes reconstructed using single image, single LiDAR sweep and single image + single LiDAR sweep together. Monocular Image Reconstruction is a feed-forward approach, where a single image is used to reconstruct the 3D shape. On the other hand, LiDAR and Image + LiDAR completion approaches iteratively optimize the 3D shape at test-time as proposed in the paper. As can be seen in the figure, shapes reconstructed using monocular image have high fidelity, pertaining to the proposed multi-stage training regime. Further, adding the test-time optimization significantly boosts performance compared to a feed-forward network, making the generated shapes register well with the ground-truth shape.


